\documentclass[conference]{IEEEtran}
\IEEEoverridecommandlockouts
\usepackage[nospace]{cite}
\usepackage{amsmath,amssymb,amsfonts}
\usepackage{graphicx}
\usepackage{textcomp}
\usepackage{xcolor}
\usepackage{subcaption}
\usepackage{bm}
\usepackage{url}
\usepackage{multirow}
\usepackage{pifont}
\usepackage{algorithm}
\usepackage{algpseudocode}
\usepackage{amssymb}
\usepackage{amsthm}
\usepackage{hyperref} 
\hypersetup{
    colorlinks=true,
    linkcolor=black,
    citecolor=black,
    urlcolor=black
}
\usepackage[flushleft]{threeparttable}

\newcommand\change[1]{\textcolor{black}{#1}}

\def\BibTeX{{\rm B\kern-.05em{\sc i\kern-.025em b}\kern-.08em
    T\kern-.1667em\lower.7ex\hbox{E}\kern-.125emX}}

\usepackage{amsmath,amsfonts,bm}

\def\eqref#1{equation~\ref{#1}}

\def\1{\bm{1}}

\DeclareMathAlphabet{\mathsfit}{\encodingdefault}{\sfdefault}{m}{sl}
\SetMathAlphabet{\mathsfit}{bold}{\encodingdefault}{\sfdefault}{bx}{n}

\begin{document}

\title{Evaluation and Optimization of Gradient Compression for Distributed Deep Learning}

\author{
\IEEEauthorblockN{Lin Zhang\IEEEauthorrefmark{2}, Longteng Zhang\IEEEauthorrefmark{3}, Shaohuai Shi\IEEEauthorrefmark{4}\IEEEauthorrefmark{1}\thanks{*Corresponding author.}, Xiaowen Chu\IEEEauthorrefmark{5}\IEEEauthorrefmark{2}\IEEEauthorrefmark{1}, Bo Li\IEEEauthorrefmark{2} \\}
\IEEEauthorblockA{\IEEEauthorrefmark{2}The Hong Kong University of Science and Technology, \IEEEauthorrefmark{3}Hong Kong Baptist University, \\
\IEEEauthorrefmark{4}Harbin Institute of Technology, Shenzhen, \IEEEauthorrefmark{5}The Hong Kong University of Science and Technology (Guangzhou)\\
lzhangbv@connect.ust.hk, ltzhang@comp.hkbu.edu.hk, shaohuais@hit.edu.cn, xwchu@ust.hk, bli@cse.ust.hk}
}

\maketitle

\begin{abstract}
To accelerate distributed training, many gradient compression methods have been proposed to alleviate the communication bottleneck in synchronous stochastic gradient descent (S-SGD), but their efficacy in real-world applications still remains unclear. In this work, we first evaluate the efficiency of three representative compression methods (quantization with Sign-SGD, sparsification with Top-k SGD, and low-rank with Power-SGD) on a 32-GPU cluster. The results show that they cannot always outperform well-optimized S-SGD or even worse due to their incompatibility with three key system optimization techniques (all-reduce, pipelining, and tensor fusion) in S-SGD. To this end, we propose a novel gradient compression method, called alternate compressed Power-SGD (ACP-SGD), which alternately compresses and communicates low-rank matrices. ACP-SGD not only significantly reduces the communication volume, but also enjoys the three system optimizations like S-SGD. Compared with Power-SGD, the optimized ACP-SGD can largely reduce the compression and communication overheads, while achieving similar model accuracy. In our experiments, ACP-SGD achieves an average of $4.06\times$ and $1.43\times$ speedups over S-SGD and Power-SGD, respectively, and it consistently outperforms other baselines across different setups (from 8 GPUs to 64 GPUs and from 1Gb/s Ethernet to 100Gb/s InfiniBand). 
\end{abstract}


\section{Introduction}\label{sec:intro}
Training deep neural networks (DNNs) with synchronous stochastic gradient descent (S-SGD) with data parallelism is one of the most popular approaches for distributed deep learning (DL)~\cite{goyal2017accurate,jia2018highly}. The S-SGD mainly consists of two phases: gradient computation and gradient aggregation. During the aggregation phase, it exchanges the locally calculated gradients through the network. With the increase of model size of DNNs (millions to billions of parameters), distributed training has caused high communication overheads, which becomes the performance bottleneck in S-SGD~\cite{peng2019generic,Zhang2020IsNT,Tang20211bitAdam}. 

To mitigate the communication overheads, on one hand, several system optimization techniques have been proposed to improve the performance of S-SGD, such as the ring all-reduce primitive~\cite{baidu2017ring}, wait-free back-propagation (WFBP)~\cite{zhang2017poseidon}, and tensor fusion~\cite{sergeev2018horovod,shi2019mg}. The ring all-reduce is known to be bandwidth optimal, so it is suitable for distributed training to transfer long messages~\cite{thakur2005optimization}. In the all-reduce based architecture, due to the deep structure of DNNs, wait-free back-propagation and tensor fusion help to better overlap communication overheads with computation overheads of different layers~\cite{shi2020quantitative,shi2021mg}. For example, in our experiments (see Fig.~\ref{fig:ablation-opt}), the optimized S-SGD (with WFBP and tensor fusion) can achieve almost $73\%$ performance improvement over the naive implementation when training a ResNet-152 model~\cite{he2016deep}. 

On the other hand, gradient compression methods have received much attention to reduce the communication volume of transferred data in different manners, including quantization~\cite{Seide20141bitSGD, wen2017terngrad,Alistarh2017QSGD,Bernstein2018signSGD,Bai2021GradientCS}, sparsification~\cite{lin2018deep,Han2020AdaptiveGS,shi2021towards,li2022near}, and low-rank decomposition~\cite{wang18atomo,vogels2019powersgd}. For example, Sign-SGD~\cite{Bernstein2018signSGD} communicates only the signs of gradients, and Top-$k$ SGD~\cite{shi2021towards} exchanges a fraction (e.g., $0.1\%$) of selected gradients. Compared to S-SGD, they are able to reduce the communication traffic by $32\times$ and even $1000\times$ times for gradient aggregation, with a negligible impact on model accuracy~\cite{lin2018deep}. 

To study the efficiency of current gradient compression methods, we choose three representative compression methods: Sign-SGD~\cite{Bernstein2018signSGD}, Top-$k$ SGD~\cite{shi2021towards}, and Power-SGD~\cite{vogels2019powersgd}, and compare their practical training performance against the well-optimized S-SGD (i.e. implemented with aforementioned system optimization techniques), in a typical data-center setting with a 32-GPU cluster connected with 10Gb/s Ethernet (10GbE). However, we find that three gradient compression methods fail to provide performance improvements over S-SGD in many cases, while they can achieve high compression ratios. For example, it is unexpected that S-SGD runs 21\%-70\% faster than compression counterparts in training ResNet-50 (see Fig.~\ref{fig:gradient-compression-time}). This surprising observation motivates us to optimize gradient compression from the system perspective. 

However, current gradient compression methods are not compatible with common system optimizations, as they do not incur either additive or non-blocking communications. To be precise, Sign-SGD and Top-$k$ SGD do not support gradient summation after quantization/sparsification, thus they are unable to utilize ring all-reduce for gradient aggregation. Meanwhile, the communication of Power-SGD will block the subsequent operations, causing trouble for wait-free back-propagation. Due to these limitations, it is non-trivial to integrate existing system optimizations into compression. 

In this work, we propose a novel gradient compression method, called alternate compressed Power-SGD (ACP-SGD)\footnote{We provide open-source code in \url{https://github.com/lzhangbv/acpsgd}.}, to enable common system optimizations like S-SGD. The idea of ACP-SGD is derived from Power-SGD, which compresses each large gradient matrix into two low-rank matrices ($P$ and $Q$) using power iteration. Unlike Power-SGD, we do not calculate and aggregate $P$ and $Q$ in one iteration, but compress the gradient into either $P$ and $Q$ alternately. In other words, ACP-SGD only needs to compress and then aggregate the gradient once in each iteration, which implies that the gradient communication operations are additive and non-blocking like S-SGD. Another benefit of ACP-SGD is to reduce half gradient compression and aggregation costs compared to Power-SGD. In addition, we apply reuse and error feedback mechanisms to improve the approximation quality and incorporate the approximation error. By doing so, ACP-SGD can achieve model accuracy on par with S-SGD. 

To optimize ACP-SGD, we use 1) ring all-reduce to aggregate $P$ or $Q$ in each iteration, 2) wait-free back-propagation to overlap these all-reduce communication tasks with gradient computation and compression tasks, and 3) tensor fusion to merge small tensors to be communicated together to reduce the start-up cost. For tensor fusion, we use compressed buffer size to determine the fusion results on $P$ and $Q$, which has shown to be adaptive to the choice of ranks (compression ratios). 

We conduct extensive experiments to validate the efficiency of our ACP-SGD with system optimizations, on 32 GPUs connected with 10GbE. The experimental results show that (1) ACP-SGD consistently outperforms S-SGD and Power-SGD in many setups, e.g., different models, batch sizes, compression ratios, and network bandwidths, (2) ACP-SGD achieves an average of $4.06\times$ and $1.43\times$ (up to $9.42\times$ and $2.11\times$) speedups over S-SGD, and Power-SGD, respectively, and (3) system optimization techniques integrated in ACP-SGD help achieve $2.14\times$ performance improvement over the naive implementation. 

\section{Background and Related Work}
In this section, we present the background and some related work of distributed S-SGD and gradient compression methods in DL training. 

\subsection{S-SGD with System Optimizations}
\begin{figure}[!t]
    \centering
    \includegraphics[width=0.85\columnwidth]{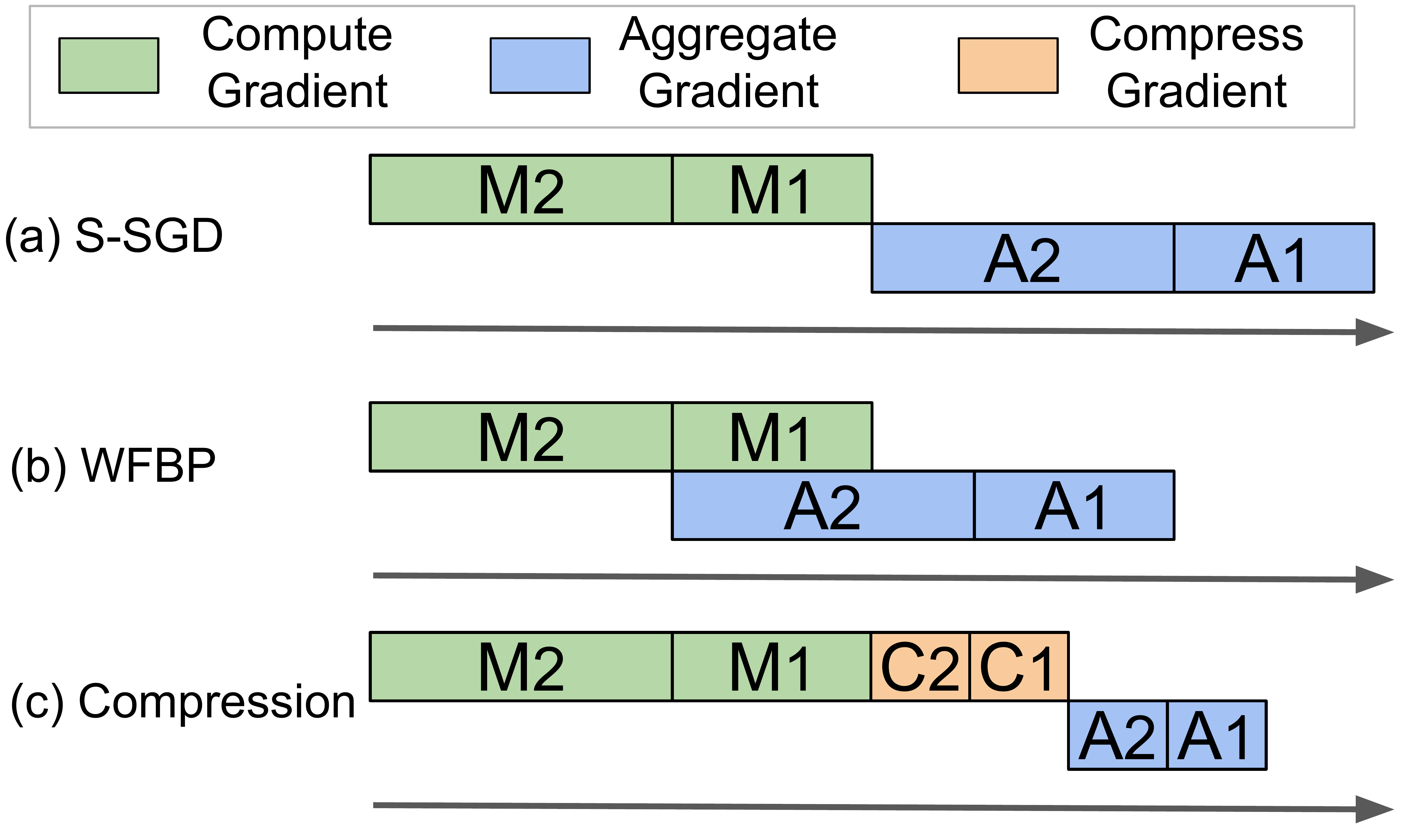}
    \caption{An illustration of how wait-free back-propagation (WFBP) and gradient compression can reduce the total iteration time for training a two-layer neural network. (a) S-SGD without any system optimization, i.e., aggregating gradients after back-propagation; (b) S-SGD with WFBP, i.e., overlapping the computing and communication tasks; and (c) gradient compression can reduce the communication traffic. }
    \label{fig:compression-example}
\end{figure}

\textbf{S-SGD. } The S-SGD is one of the most popular algorithms used to accelerate DNN training with data parallelism~\cite{goyal2017accurate,jia2018highly}. Each iteration of S-SGD consists of two main phases: gradient computation and gradient aggregation. During the gradient computation phase, the local gradient at each worker is computed via feed-forward and back-propagation (FF\&BP) process, followed by the gradient aggregation. As shown in Fig.~\ref{fig:compression-example}(a), local gradients are synchronously communicated (summed up) among all workers, so that each worker has the aggregated global gradient for the model update. Due to the large size of deep models, millions to billions of parameters are communicated among workers, leading to significant performance bottlenecks~\cite{peng2019generic,Zhang2020IsNT,Tang20211bitAdam}.

\textbf{System Optimizations. } To alleviate the communication overheads, on one hand, many system level optimization techniques~\cite{shi2020quantitative} have been proposed to improve the training efficiency of S-SGD. In this paper, we introduce three commonly used system advances : ring all-reduce primitive~\cite{baidu2017ring}, wait-free back-propagation~\cite{zhang2017poseidon}, and tensor fusion~\cite{sergeev2018horovod,Li2020PyTorchDDP}. 

\subsubsection{Ring All-reduce} by building a ring-based topology for simultaneously communicating gradient chunks between any neighbor workers, ring all-reduce enjoys a very high bandwidth, and its communication complexity is linear to the number of parameters, no matter of the cluster size (as shown in Table~\ref{tab:comm-complex}). On top of efficient implementations such as NCCL, ring all-reduce primitive has successfully become the major communication protocol for high performance distributed training in recent years~\cite{you2018imagenet,you2020large}. 

\subsubsection{Wait-free Back-propagation} due to the deep structure of DNNs, the gradients are calculated layer-wisely during the back-propagation. Instead of waiting for the completion of back-propagation to calculate all gradients, existing training frameworks support wait-free back-propagation (WFBP)~\cite{zhang2017poseidon}, where gradient communication can start immediately when the gradient is ready. This enables gradient communication to be overlapped with gradient computation tasks of previous layers, hiding the time spent in communication. For example, in Fig.~\ref{fig:compression-example}(b), when the gradient $M_2$ of layer $2$ is ready, it starts to aggregate $M_2$ immediately (i.e., $A_2$), so that communication task of $A_2$ can be overlapped with the gradient computation task of $M_1$ of layer $1$. 

\subsubsection{Tensor Fusion} naive WFBP by calling ring all-reduce operations layer-wisely can however lead to heavy communication overheads. This is because each all-reduce also has a start-up cost, which is unfortunately linear to the number of workers~\cite{thakur2005optimization}, making all-reducing small tensors separately very inefficient. For instance, on our 10GbE platform, all-reducing two $32$KB tensors takes about $2.0$ms, while all-reducing one $64$KB tensor only requires $1.2$ms. Based on this observation, existing training frameworks~\cite{sergeev2018horovod,Li2020PyTorchDDP} have equipped WFBP with tensor fusion (TF)~\cite{shi2019mg,shi2021mg,shi2021exploiting}, to merge small gradient tensors of nearby layers into one large tensor. With TF, fewer all-reduce operations are required to aggregate these merged large gradient tensors, which amortizes the start-up costs. 

In this paper, we implement S-SGD with aforementioned system optimization techniques as the baseline. 

\subsection{Gradient Compression Methods}
Another line of work has proposed gradient compression methods to mitigate gradient communication costs~\cite{Xu2021Grace}. Gradient compression methods aim to reduce the communication traffic by compressing the gradients, as demonstrated in Fig~\ref{fig:compression-example}(c), in the following three categories. 

\subsubsection{Quantization} Quantization methods reduce the number of bits of each element of the gradients. 1-bit SGD~\cite{Seide20141bitSGD}, Sign-SGD~\cite{Bernstein2018signSGD,Karimireddy2019ErrorFF}, 1-bit Adam~\cite{Tang20211bitAdam} quantize each float32 element of the gradient to 1-bit sign, by mapping the negative components to $-1$ and the others to $+1$. TernGrad~\cite{wen2017terngrad} and QSGD~\cite{Alistarh2017QSGD} quantize each element into three $(-1, 0, 1)$ or more values via randomized rounding. However, quantization can at most reduce the communication volume by 32$\times$. 

\subsubsection{Sparsification} To reduce communication traffic in a more aggressive way, sparsification methods select only a small subset of gradient elements (e.g., $0.1\%$), resulting in a sparse communication~\cite{lin2018deep,Han2020AdaptiveGS,shi2021towards,m2021efficient}. Let $k$ be the number of selected gradients, Random-$k$ and Top-$k$ are two representatives to choose the $k$ random coordinates and $k$ largest coordinates (in magnitude), respectively, where Top-$k$ tends to achieve better convergence performance than Random-$k$ in practise~\cite{Stich2018SparsifiedSGD}. For Top-$k$ SGD, one needs to communicate selected gradients and their indices~\cite{lin2018deep,shi2019distributed,renggli2019sparcml}. 

\subsubsection{Low-rank Decomposition} Given a large gradient matrix $M \in \mathbb{R}^{n \times m}$, low-rank decomposition methods factorize $M$ into two low-rank matrices $M \approx P Q^T$, where $P \in \mathbb{R}^{n \times r}$ and $Q \in \mathbb{R}^{m \times r}$. As the rank $r$ is smaller than $n$ and $m$, communicating $P$ and $Q$ is more efficient than communicating $M$ by a factor of $(nm)/(nr+mr)$. ATOMO~\cite{wang18atomo} uses compute-expensive SVD to achieve low-rank decomposition, and Power-SGD~\cite{vogels2019powersgd} utilizes power iteration to calculate low-rank matrices with much cheaper decomposition costs, which makes Power-SGD relatively practical in real-world applications~\cite{ramesh2021zero}. The procedure of Power-SGD is given in Algorithm~\ref{algo:power-sgd}. It involves two matrix multiplications, one orthogonalization, and two all-reduce operations to compute and aggregate $P$ and $Q$. 

While gradient compression methods can dramatically reduce the communication traffic for gradient aggregation, they are non-trivial to be compatible with existing system optimization techniques that are oft-used in S-SGD, which adversely affects the practical scalability of gradient compression methods, as discussed in the next section. 

\textbf{Other Related Work.} \cite{Xu2021Grace,AgarwalWVP22utility} present quantitative evaluation of gradient compression methods, observing the unsatisfying performance of current gradient compression methods. To improve their scaling efficiency, HiPress~\cite{Bai2021GradientCS} splits gradient aggregation into composable and pipelined computing and communication primitives, and ByteComp~\cite{Wang2022ByteComp} attempts to search the optimal compression strategy with a decision tree. However, these works require much effort to develop new tools, without maximizing the potential of combining current system optimizations into gradient compression. 
\change{For accelerating Top-$k$ SGD, sparse all-reduce algorithms~\cite{shi2019distributed,li2022near} and statistical-based top-k selection algorithms~\cite{shi2019understanding,m2021efficient} are specifically developed to alleviate the sparse communication and compression overheads, respectively, but unfortunately they are not compatible to each other~\cite{li2022near}.}  


\section{Characterizing Gradient Compression}\label{sec:benckmark-exp}
In this section, we characterize the performance of three representative gradient compression methods, including Sign-SGD, Top-$k$ SGD, and Power-SGD. 

\subsection{Experimental Settings}\label{sec:benckmark-exp-setting}
\textbf{Methods:} we compare the performance of different gradient compression methods with S-SGD which is well optimized atop off-the-shelf implementation of PyTorch-DDP~\cite{Li2020PyTorchDDP}. Following the prior work~\cite{AgarwalWVP22utility}, we choose three most scalable gradient compression methods in each category: Sign-SGD with majority vote~\cite{Bernstein2018signSGD}, Top-$k$ SGD with multiple sampling~\cite{shi2021towards}~\footnote{\change{The exact Top-$k$ selection is very computationally inefficient in GPUs, instead, multiple sampling uses binary
search to find a close top-k threshold. }}, and Power-SGD~\cite{vogels2019powersgd}. For Power-SGD, we use the all-reduce collective, while for Sign-SGD and Top-$k$ SGD, we use the all-gather collective~\cite{renggli2019sparcml}. The gradients are packed together to be compressed and communicated for better performance~\cite{AgarwalWVP22utility}. 

\begin{table}[!ht]
    \centering
    \begin{tabular}{|c|c|c|c|c|}
        \hline
        Model & \#Param. & Sign-SGD & Top-$k$ SGD & Power-SGD \\\hline\hline
        ResNet-50 & 25.6 & 32$\times$ & 1000$\times$ & 67$\times$ \textcolor{gray}{(r=4)} \\\hline
        ResNet-152 & 60.2 & 32$\times$ & 1000$\times$ & 53$\times$ \textcolor{gray}{(r=4)} \\\hline
        BERT-Base & 110.1 & 32$\times$ & 1000$\times$ & 16$\times$ \textcolor{gray}{(r=32)} \\\hline
        BERT-Large & 336.2 & 32$\times$ & 1000$\times$ & 21$\times$ \textcolor{gray}{(r=32)} \\\hline
    \end{tabular}
    \caption{Model statistics and compression ratios of different gradient compression methods. \#Param. denotes the number of model parameters (in million). $r$ is the rank of Power-SGD. } 
    \label{tab:comp-ratio}
\end{table}

\textbf{Models:} following the work~\cite{AgarwalWVP22utility}, we choose four popular DNN models: ResNet-50~\cite{he2016deep}, ResNet-152~\cite{he2016deep}, BERT-Base~\cite{devlin2019bert}, and BERT-Large~\cite{devlin2019bert}, with per-GPU batch size of 64, 32, 32, and 8, respectively. The input image size is set as $3 \times 224 \times 224$ for ResNets, and the input sequence length is set as $64$ for BERTs. The model statistics and compression ratios are given in Table~\ref{tab:comp-ratio}. We are optimistic to use $1000\times$ compression ratio in Top-$k$ SGD. For Power-SGD, we use a rank $r=4$ for ResNets, and a higher rank $r=32$ for BERTs to achieve competitive results as suggested in~\cite{vogels2019powersgd}. 

\textbf{Testbed:} we conduct our experiments on a 32-GPU cluster of 8 nodes. Each node has 4 Nvidia RTX2080Ti GPUs (11GB RAM) connected by PCIe3.0x16. The interconnect between nodes is 10Gb/s Ethernet (10GbE). The common software includes PyTorch-1.12.1, CUDA-11.3, cuDNN-8.3.2, and NCCL-2.10.3. 

\textbf{Metric:} we use the metric of average iteration time in running 120 iterations (exclude the first 20 iterations). Each gradient compression method mainly consists of gradient computation (i.e., FF\&BP), gradient compression and decompression, and gradient communication costs. For the communication cost, we only measure its non-overlapped overhead. 

\subsection{Performance Comparison}
\begin{figure}[!ht]
    \centering
    \includegraphics[width=0.85\columnwidth]{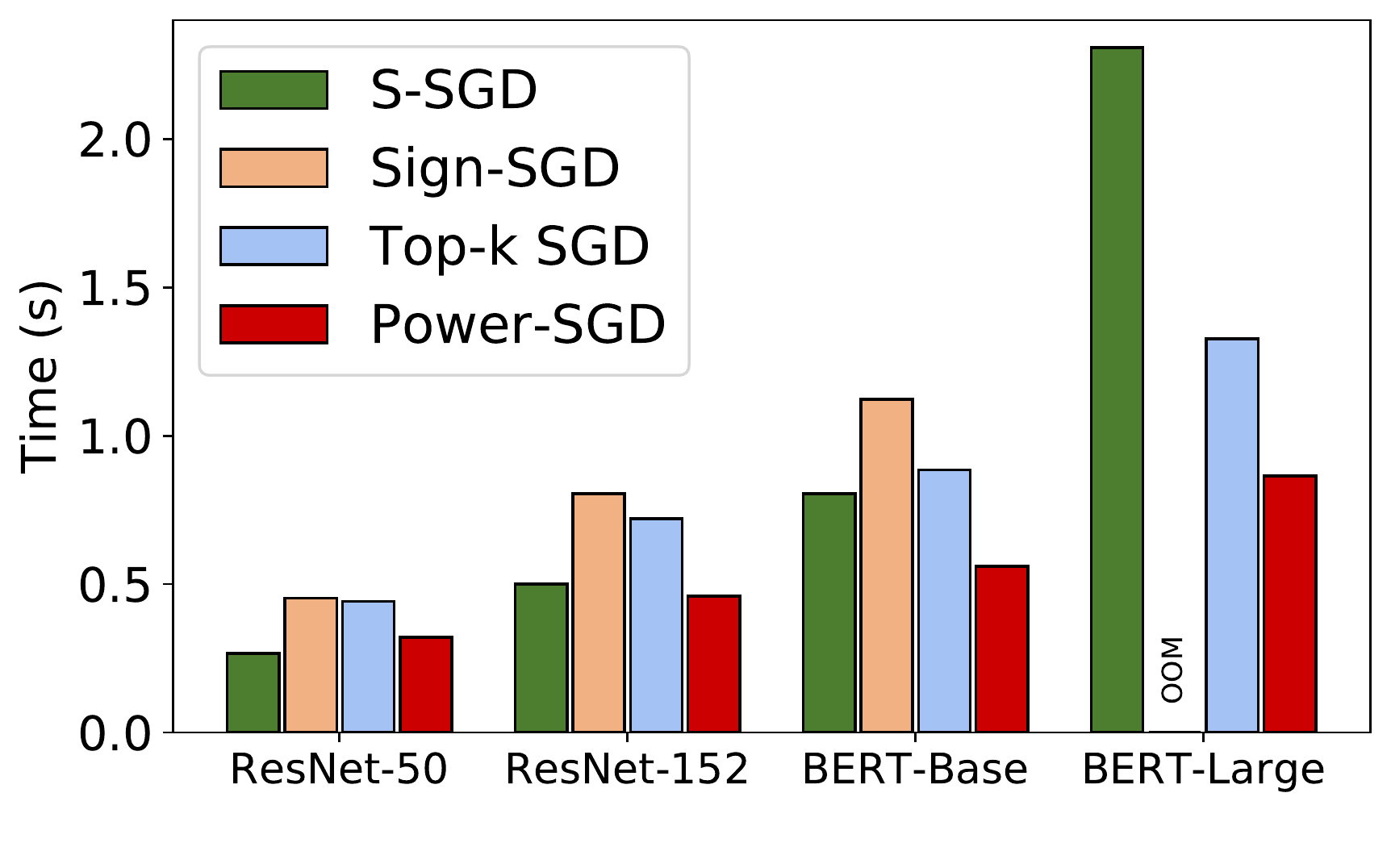}
    \caption{Average iteration wall-clock time (in seconds) comparison of different gradient compression methods. We do not report error bars as the std is generally very small ($\le$10ms). }
    \label{fig:gradient-compression-time}
\end{figure}

We report iteration time of S-SGD, Sign-SGD, Top-$k$ SGD, and Power-SGD in Fig.~\ref{fig:gradient-compression-time}. It shows that Sign-SGD and Top-$k$ SGD usually perform worse than S-SGD, while they are able to compress the gradients by $32\times$ and $1000\times$ times, respectively. For example, Sign-SGD and Top-$k$ SGD take $1.70\times$ and $1.66\times$ higher iteration time than S-SGD for training a ResNet-50 model. For training the largest BERT-Large model, the Top-$k$ SGD runs faster than S-SGD, while Sign-SGD runs out of memory due to its increased memory requirement. Among three gradient compression methods, Power-SGD has achieved the best performance over all four models. However, Power-SGD outperforms S-SGD only on large models (BERT-Base and BERT-Large), and it performs worse or closely than S-SGD on small models (ResNet-50 and ResNet-152). Overall, it is disappointing to find that three gradient compression methods, with $16\times$ to $1000\times$ high compression ratios, fail to achieve speedup over S-SGD in many cases on 10GbE bandwidth, which is very common in public cloud clusters~\cite{cho2019blueconnect,shi2021towards}. 

\begin{figure}[!ht]
    \centering
    \includegraphics[width=0.85\columnwidth]{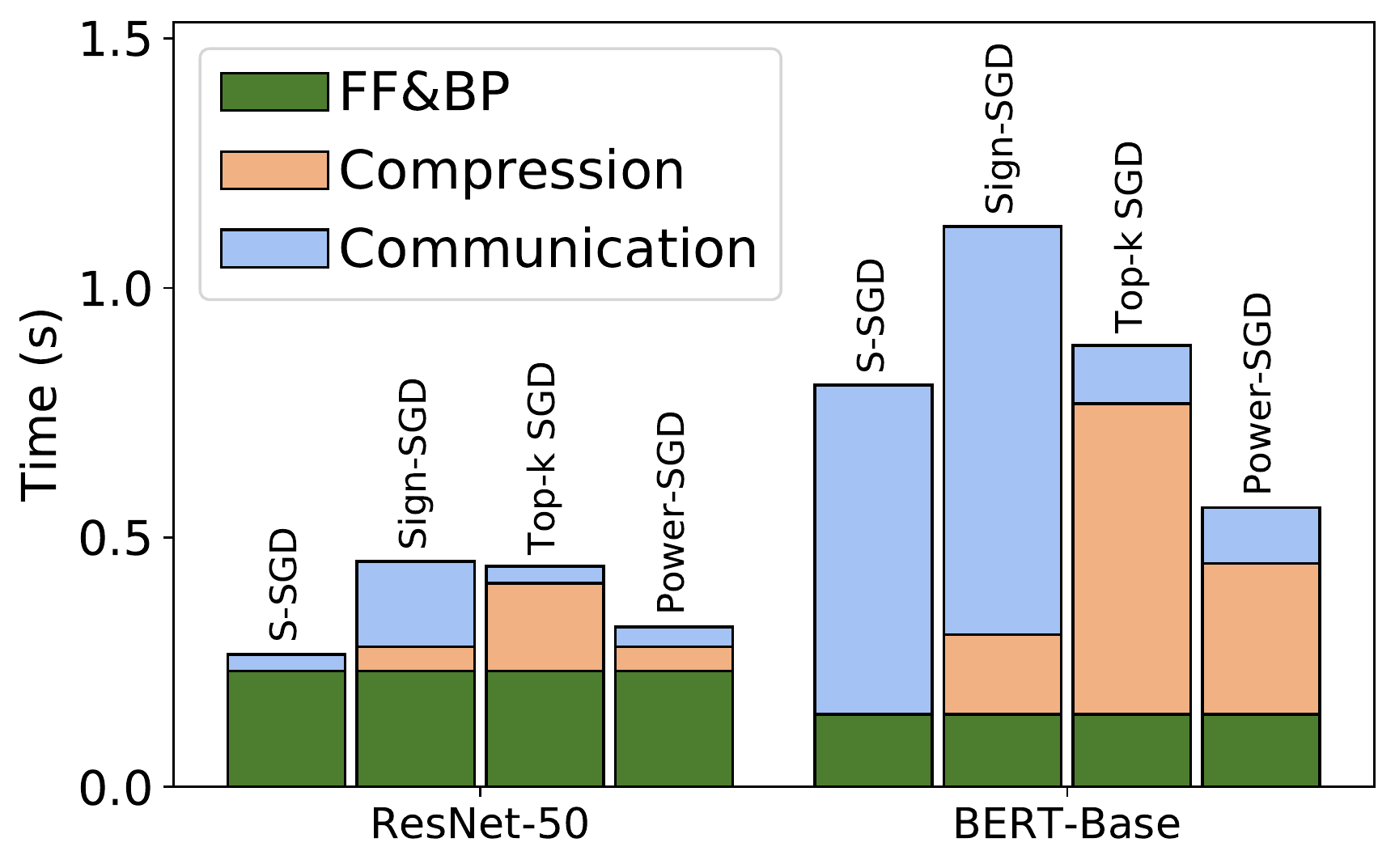}
    \caption{Time breakdowns of gradient compression methods on ResNet-50 and BERT-Base models. }
    \label{fig:gradient-compression-breakdown}
\end{figure}

To understand the performance bottleneck, we dive into time breakdowns of gradient compression methods. Specifically, we measure the FF\&BP computation time, compression and decompress time, and non-overlapped communication time. The time breakdowns of ResNet-50 and BERT-Base are shown in Fig.~\ref{fig:gradient-compression-breakdown}. The results of ResNet-152 and BERT-Large are similar to ResNet-50 and BERT-Base, respectively. 

First, the communication overhead of S-SGD on BERT-Base is much larger than that on ResNet-50, because BERT-Base has more parameters, as well as a higher communication-to-computation ratio than ResNet-50. Thus, the communication overhead cannot be well hidden on BERT-Base, leaving more speedup space for gradient compression methods. Second, Sign-SGD has cheaper compression cost than Top-$k$ SGD and Power-SGD, however, its communication cost is even higher than uncompressed S-SGD. We contribute it to the fact that Sign-SGD has at most $32\times$ compression ratio, but it requires all-gather for communication, which is less efficient than all-reduce used in S-SGD. On the other hand, even though Top-$k$ SGD uses all-gather for communication, its communication overhead is very low due to its up to $1000\times$ compression ratio. However, Top-$k$ SGD suffers from the computation bottleneck of Top-$k$ operations, for example, its takes $4\times$ compression time to achieve $7\times$ communication speedup than Sign-SGD when training BERT-Base. \change{It is partly because our multiple sampling top-k selection implemented atop PyTorch is less efficient than its highly-optimized CUDA version~\cite{shi2021towards}, which however is not publicly available. } Third, Power-SGD achieves competitive communication performance by using all-reduce, while having a mild compression cost. For large model BERT-Base, Power-SGD is $1.4\times$ faster than S-SGD, showing it is one of the most promising gradient compression methods to outperform S-SGD. 

In summary, existing gradient compression methods require non-negligible compression and/or communication overheads, resulting in poor scalability. It is worth noting that the S-SGD baseline has been well optimized in the system-level, while gradient compression methods are not. We will discuss the limitations of current gradient compression methods to integrate these techniques. 

\subsection{Limitations of Gradient Compression}
For S-SGD, the ring all-reduce primitive, wait-free back-propagation, and tensor fusion optimization techniques have been widely applied to accelerate distributed training, such as in Horovod~\cite{sergeev2018horovod} and PyTorch-DDP~\cite{Li2020PyTorchDDP}. However, these techniques are not readily applied into aforementioned gradient compression methods. 

We first summarize two good properties of S-SGD to enable these system optimization techniques as follows: 
\begin{itemize}
    \item \textit{additive communication:} the gradient communication operation in each DNN layer is to sum up the local gradient tensor from each workers. The gradient addition enables efficient aggregation with ring all-reduce primitive to distribute the reduced gradient result to all workers. 
    \item \textit{non-blocking communication:} the gradient communication operation in each DNN layer is independent of backward operations, without blocking the subsequent back-propagation. It enables wait-free back-propagation and tensor fusion techniques very friendly to overlap communication with computation tasks. 
\end{itemize}

However, these two properties are not both satisfied in current gradient compression methods. For Sign-SGD and Top-$k$ SGD, each DNN layer requires gradient calculation and gradient quantitation/sparsification tasks, followed by communicating the compressed gradient, so the communication will not block the subsequent back-propagation~\cite{shi2020communication}. However, two quantized gradients cannot be simply added, e.g., the result of adding two $+1$ values is out of the range in Sign-SGD. Two sparsified gradient tensors are not additive as well, because their selected gradient elements (i.e., Top-$k$ elements) have different coordinates. As a compromise, Top-$k$ SGD and Sign-SGD typically utilize the all-gather primitive~\cite{renggli2019sparcml} to collect quantized and/or sparsified gradient results. Specifically, the communication complexity comparison is given in Table~\ref{tab:comm-complex}, showing that the communication complexity of Sign-SGD and Top-$k$ SGD (with all-gather) is linear to the number of workers, which is system inefficient compared to S-SGD (with all-reduce)~\cite{shi2021towards,AgarwalWVP22utility}. Sign-SGD lacks compatibility with all-reduce, as studied in Fig.~\ref{fig:gradient-compression-breakdown}, so its actual communication overhead (with $32\times$ compression ratio) is however $24\%$ higher than S-SGD when training BERT-Base. 

\begin{table}[!ht]
    \centering
    \addtolength{\tabcolsep}{-1.5pt}
    \begin{tabular}{|c|c|c|c|c|}
        \hline
        \textbf{Complexity} & S-SGD & Sign-SGD & Top-$k$ SGD & Power-SGD  \\ \hline\hline
        \textbf{Compress} & -- & $O(N)$ & $O(k \log N)$ & $O(Nr)$  \\ \hline
        \textbf{Communicate} & $\frac{2(p-1)}{p} N$ & $(p-1) \frac{N}{32}$ & $(p-1) 2k$ & $\frac{2(p-1)}{p} N_c$  \\ \hline
    \end{tabular}
    \caption{Compress and communicate complexity comparison among different algorithms, where $p$ is the number of workers, $N$ is the number of uncompressed gradients, $k$ is the number of selected gradients, and $N_c$ is the number of compressed gradients in Power-SGD (with the rank of $r$).}  
    \label{tab:comm-complex}
\end{table}

\begin{figure}[!t]
    \centering
    \includegraphics[width=0.8\columnwidth]{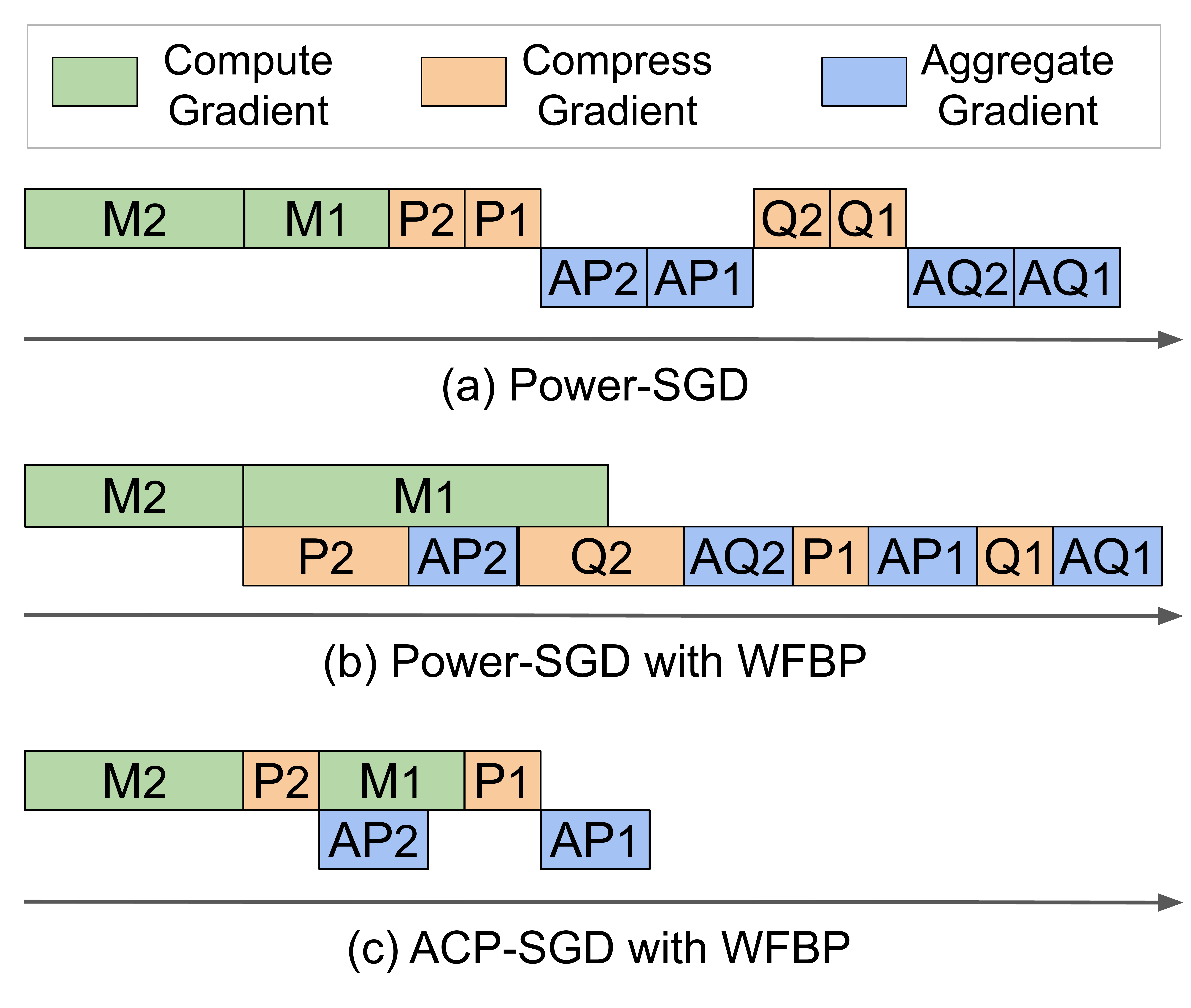}
    \caption{An illustration of how wait-free back-propagation (WFBP) can harm the performance of Power-SGD, while improving the performance of ACP-SGD. (a) Power-SGD computes and aggregates $P$ and $Q$ after back-propagation; (b) Power-SGD with WFBP overlaps gradient compression with gradient computation, affecting the back-propagation time (e.g., slowdown of $M_1$); (c) ACP-SGD with WFBP overlaps aggregation tasks (e.g., $AP_2$) with computing tasks. }
    \label{fig:acp-sgd}
\end{figure}

On the other hand, the compressed gradients of Power-SGD are still dense matrices, which allows Power-SGD to use efficient ring all-reduce primitive to aggregate these smaller matrices (i.e., $N_c \ll N$ as shown in Table~\ref{tab:comm-complex}). However, in Power-SGD, each DNN layer needs to calculate the gradient $M$ and low-rank $P$, followed by aggregating $P$, and then it needs to calculate $Q$ based on the result of aggregated $P$, followed by aggregating $Q$, as shown in Fig.~\ref{fig:acp-sgd}(a). In other words, the communication of aggregating $P$ will block the subsequent operations of computing and aggregating $Q$, and causes trouble for overlapping communication tasks (i.e., aggregating $P$ and $Q$) with back-propagation. As demonstrated in Fig.~\ref{fig:acp-sgd}(b), Power-SGD with WFBP will overlap the whole gradient compression and communication process with back-propagation. By doing so, gradient compression and communication tasks (e.g., $P_2$, $AP_2$, $Q_2$) are performed in parallel with gradient computation tasks (e.g., $M_1$). However, as gradient compression and gradient computation are both compute intensive tasks (see Table~\ref{tab:comm-complex},  the compress complexity of Power-SGD), they will compete for compute resources on the GPU, leading to performance slowdown~\cite{AgarwalWVP22utility}, i.e., affecting computing tasks of $M_1$ and $P_2$ in Fig.~\ref{fig:acp-sgd}(b). For example, Power-SGD with WFBP causes an overall of $13\%$ slowdown than Power-SGD without WFBP, when training ResNet-50 on one GPU (with only computation tasks). 

Motivated by these limitations, we propose a novel gradient compression algorithm that can satisfy two good properties of S-SGD, providing opportunities for system optimizations. 

\section{Optimizing Gradient Compression} 
In this section, we present our proposed gradient compression algorithm, namely ACP-SGD, equipped with system optimization techniques, to improve the throughput of distributed DL training. ACP-SGD is an alternate compressed Power-SGD that enjoys the good property to use all-reduce to aggregate compressed gradients and it is able to utilize WFBP and TF as like in S-SGD to further improve its performance. 

\subsection{ACP-SGD: Alternate Compressed Power-SGD}
\textbf{Power-SGD.} The Power-SGD algorithm~\cite{vogels2019powersgd} uses a single step of power iteration to decompose the large gradient matrix into two low-rank matrices as $M \approx P Q^T$. As shown in Algorithm~\ref{algo:power-sgd}, it requires one right multiplication, one left multiplication, and an orthogonalization operation to compute low-rank matrices $P$ and $Q$. Besides, it calls two all-reduce operations to aggregate $P$ and $Q$, respectively. 

In each power iteration, the previous result at the last step (i.e., $Q_{t-1}$) will be re-used to update the low-rank matrices at the current step (i.e., $P_t$ and $Q_t$). This query reuse trick helps Power-SGD approximate the stochastic gradient matrix at each step more accurately~\cite{vogels2019powersgd}. In the very beginning, $Q$ is initialized from an i.i.d. standard normal distribution. 

However, Power-SGD involves interleaving computing and communication tasks (i.e., compute $P \rightarrow$  aggregate $P \rightarrow$ compute $Q \rightarrow$ aggregate $Q$) in each DNN layer. That is, the communication tasks will block the subsequent back-propagation operations. And if we simply pipeline the gradient compression and aggregation tasks with gradient computing tasks, gradient compression and gradient computation will compete for compute resources on the GPU, leading to the performance interference and slowdown~\cite{AgarwalWVP22utility}. 

\begin{algorithm}[!t]
	\caption{Power-SGD vs. ACP-SGD Compression}\label{algo:power-sgd}
	\begin{algorithmic}[1]
		\small
		\State \textbf{input}: current gradient matrix $M_t \in \mathbb{R}^{n \times m}$, previous low rank matrices $P_{t-1} \in \mathbb{R}^{n \times r}$ and $Q_{t-1} \in \mathbb{R}^{m \times r}$, and $r$ is the rank. 
		
        \Function{Power-SGD compression}{$M_t$, $Q_{t-1}$}
        \State $P_{t} \gets M_{t} Q_{t-1}$ \Comment{Compute $P$}
        \State $P_{t} \gets \text{All-Reduce}(P_t)$ \Comment{Aggregate $P$}
        \State $P_{t} \gets \text{Orthogonalize}(P_t)$
        \State $Q_{t} \gets M_{t}^T P_t$ \Comment{Compute $Q$}
        \State $Q_{t} \gets \text{All-Reduce}(Q_t)$ \Comment{Aggregate $Q$}
        \State \Return $P_t Q_t^T$
        \EndFunction
        
        \Function{ACP-SGD compression}{$M_t$, $P_{t-1}$, $Q_{t-1}$}
        \If{$t$ is odd}
        \State $Q_{t} \gets \text{Orthogonalize}(Q_{t-1})$
        \State $P_{t} \gets M_{t} Q_{t}$ \Comment{Compute $P$}
        \State $P_{t} \gets \text{All-Reduce}(P_t)$ \Comment{Aggregate $P$}
        \Else
        \State $P_{t} \gets \text{Orthogonalize}(P_{t-1})$
        \State $Q_{t} \gets M_{t}^T P_t$ \Comment{Compute $Q$}
        \State $Q_{t} \gets \text{All-Reduce}(Q_t)$ \Comment{Aggregate $Q$}
        \EndIf
        \State \Return $P_t Q_t^T$
        \EndFunction
	\end{algorithmic}
\end{algorithm}

\textbf{Alternate compression.} To avoid this, we propose an alternate compressed Power-SGD (ACP-SGD) algorithm (Algorithm~\ref{algo:power-sgd}), which compresses the gradient into $P$ and $Q$ alternately. It only requires computing $P$ (or $Q$), followed by aggregating $P$ (or $Q$) in each iteration. By doing so, the communication of ACP-SGD is not blocking anymore, leading to the benefits of overlapping computing and communication tasks like S-SGD. We defer the details of system optimization to the next subsection. 

In each power iteration, ACP-SGD reuses the previous low-rank results ($P_{t-1}$ or $Q_{t-1}$) to approximate the stochastic gradient. Consider two consecutive iterations with gradients $M_{t-1}$ and $M_t$, alternate compression is equal to performing a complete step of power iteration, except that it computes $P$ with $M_{t-1}$ but computes $Q$ with $M_{t}$. Under a small update stepsize, one can expect that $M_{t}$ is close to $M_{t-1}$~\cite{vogels2019powersgd}. Thus, query reuse is helpful to ensure the approximation quality. 

In addition, another benefit of ACP-SGD is that it can halve the gradient compression and communication costs compared to Power-SGD. That is, ACP-SGD performs one orthogonalization and one matrix multiplication with the computation complexity of $O(\frac{n+m}{2}r^2 + nmr)$, and one all-reduce primitive with the communication volume of $O(\frac{n+m}{2}r)$. 

\textbf{Error-feedback.} The error-feedback mechanism computes the difference between one worker's gradient and the compressed gradient (i.e., error), and adds it back to the next gradient (i.e., feedback)~\cite{Seide20141bitSGD,Alistarh2017QSGD,karimireddy2019error,vogels2019powersgd}. It has been shown to be crucial to improve the convergence performance of biased compression methods (i.e., compressed gradient is not equal to the original one in expectation), such as Sign-SGD, Top-$k$ SGD, and Power-SGD. Our ACP-SGD compression (derived from Power-SGD) is biased, therefore, it requires error-feedback to achieve good performance as shown in Algorithm~\ref{algo:acp-sgd-ef}. 

Take computing and aggregating $P$ as an example, we apply the error-feedback into ACP-SGD as follows: 1) incorporate the previous local error $E_{t-1}$ into the local gradient $M_t$ before compression, i.e., $P_t \gets (M_t + E_{t-1})Q_t$, and 2) update the local error via $E_t \gets M_t + E_{t-1} - P_tQ_t^T$ before aggregation. ACP-SGD with error-feedback performs the layer-wise computation (including compute $M_t$, orthogonalize $Q_{t-1}$, compute $P_t$, and update $E_t$), followed by communication of aggregating $P_t$, whose communication operation is still non-blocking. In the very beginning, $E_0$ is initialized as zeros, and low-rank matrices $P_0$ and $Q_0$ are initialized randomly from standard normal distribution. 

\begin{algorithm}[!t]
	\caption{ACP-SGD with Error-feedback (EF)}\label{algo:acp-sgd-ef}
	\begin{algorithmic}[1]
		\small
		\State \textbf{input}: current gradient matrix $M_t \in \mathbb{R}^{n \times m}$, previous low rank matrices $P_{t-1} \in \mathbb{R}^{n \times r}$ and $Q_{t-1} \in \mathbb{R}^{m \times r}$, previous error-feedback matrix $E_{t-1} \in \mathbb{R}^{n \times m}$, and rank $r$. 
        \Function{ACP-SGD with EF}{$M_t$, $P_{t-1}$, $Q_{t-1}$, $E_{t-1}$}
        \If{$t$ is odd}
        \State $Q_{t} \gets \text{Orthogonalize}(Q_{t-1})$
        \State $P_{t} \gets (M_{t}+ E_{t-1}) Q_{t}$ \Comment{Compute $P$}
        \State $E_{t} \gets M_{t}+ E_{t-1} - P_{t}Q_{t}^T$ \Comment{Update $E$}
        \State $P_{t} \gets \text{All-Reduce}(P_t)$ \Comment{Aggregate $P$}
        \Else
        \State $P_{t} \gets \text{Orthogonalize}(P_{t-1})$
        \State $Q_{t} \gets (M_{t} + E_{t-1})^T P_t$ \Comment{Compute $Q$}
        \State $E_{t} \gets M_{t}+ E_{t-1} - P_{t}Q_{t}^T$ \Comment{Update $E$}
        \State $Q_{t} \gets \text{All-Reduce}(Q_t)$ \Comment{Aggregate $Q$}
        \EndIf
        \State \Return $P_t Q_t^T$
        \EndFunction
	\end{algorithmic}
\end{algorithm}

\subsection{System Optimization of ACP-SGD}
\textbf{Wait-free Back-propagation} (WFBP). Apart from using the ring all-reduce primitive for aggregating compressed tensors, the good property of non-blocking gradient communication allows ACP-SGD to apply WFBP to overlap communication tasks with computing tasks. As shown in Fig.~\ref{fig:acp-sgd}(c), one can aggregate the compressed tensor of layer 2 (i.e., $AP_2$) immediately after the tasks of calculating and compressing the gradient (i.e., $M_2$ and $P_2$) have been finished. By doing WFBP, the communication task of $AP_2$ can be overlapped with the computing tasks of layer 1 (i.e., $M_1$). In the next iteration, WFBP can be equally applied to overlap the computing and communication tasks for $Q$. Meanwhile, ACP-SGD with WFBP only overlap all-reduce operations that are communication intensive and do not compete for the compute resources. 

\begin{figure}[!ht]
    \centering
    \includegraphics[width=0.45\linewidth]{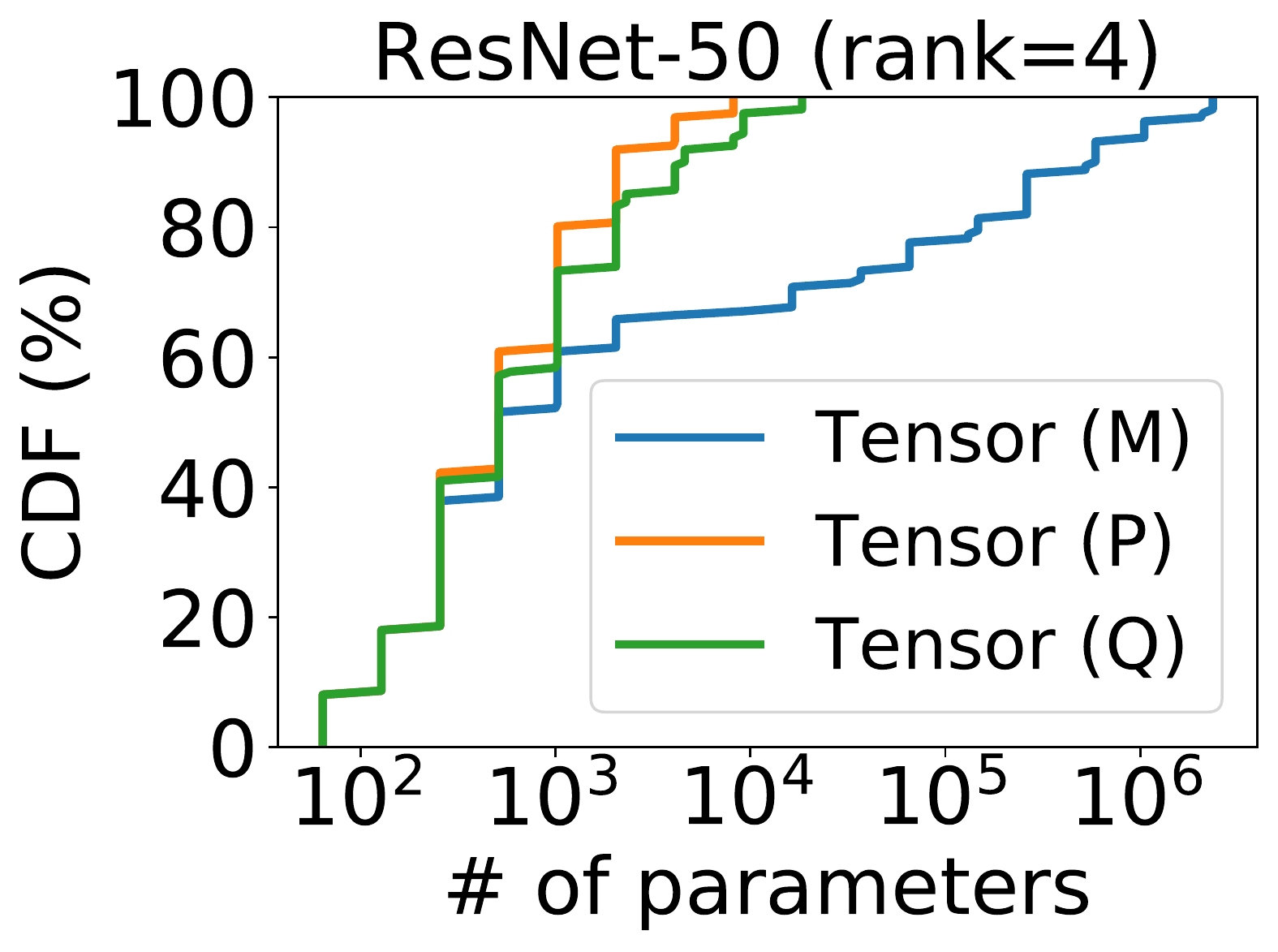}
    \includegraphics[width=0.45\linewidth]{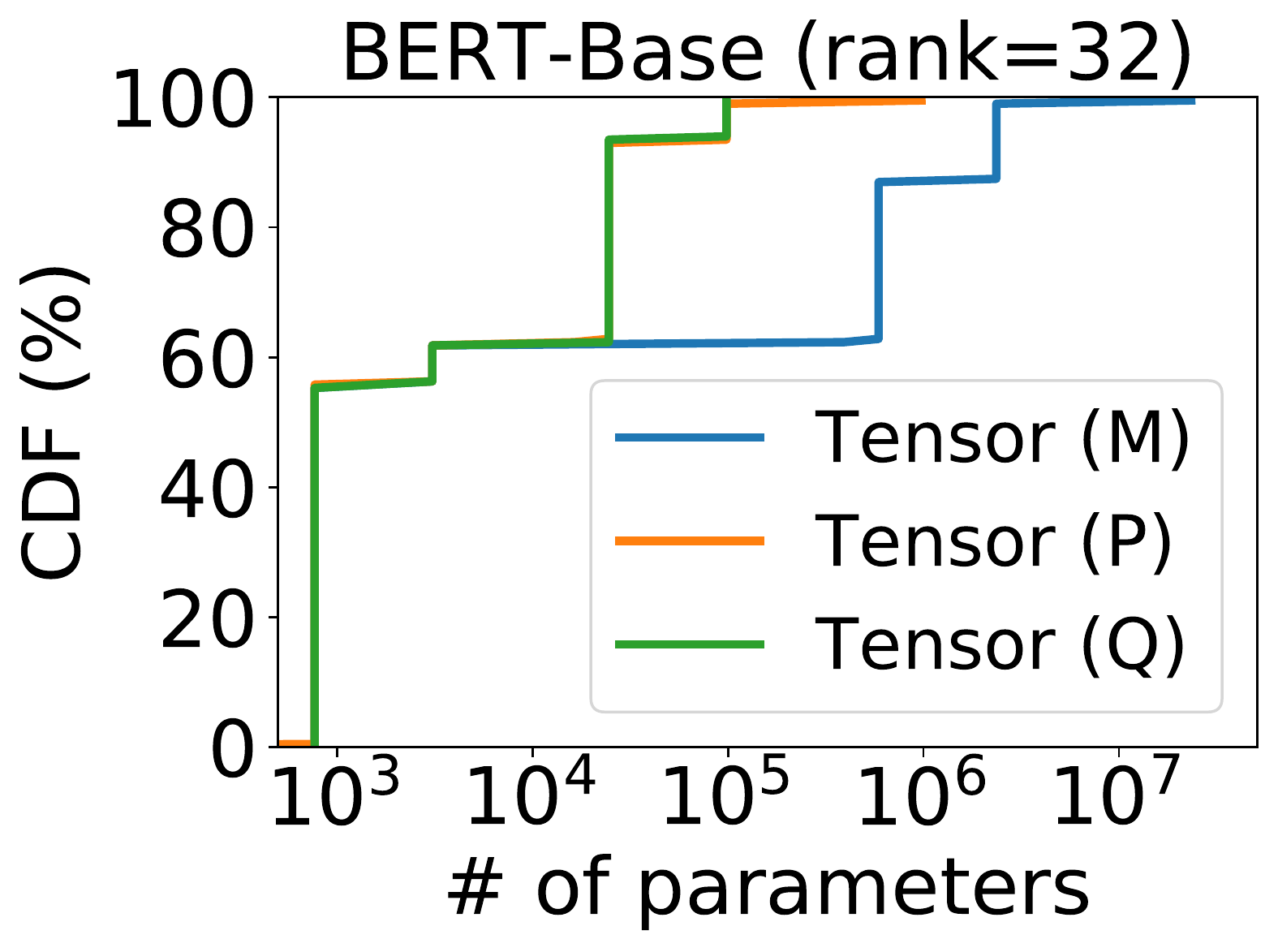}
    \caption{CDF of the number of parameters of uncompressed tensors (M) and compressed tensors (P and Q) in ACP-SGD. }
    \label{fig:cdf-param}
\end{figure}

\textbf{Tensor Fusion} (TF). If the compressed tensors $P$ and $Q$ are layer-wisely aggregated via ring all-reduce operations, the communication overheads are easily dominated by start-up costs~\cite{shi2019mg}. This is because the compressed tensors in ACP-SGD are much smaller than uncompressed tensors in S-SGD. For example, in Fig.~\ref{fig:cdf-param}, we report the cumulative distribution functions (CDFs) of the number of parameters in two models. After low-rank decomposing the gradients of ResNet-50 (or BERT-Base), we observe that there is a $30\%$ increase in the proportion of small tensors having less than $10^4$ (or $10^5$) parameters. 

Thus, all-reducing these small and compressed tensors separately can be very communication inefficient. To reduce start-up costs, one shall apply the TF technique to merge several small tensors to be communicated together via one all-reduce operation. For example, on our 10GbE platform, all-reducing uncompressed gradients of ResNet-50 takes $243$ms, while all-reducing them together takes $169$ms (with $1.4\times$ speedup). However, in ACP-SGD, all-reducing compressed tensors separately takes $55.9$ms, and all-reducing them together only takes $2.3$ms (with $24.3\times$ speedup). Although TF is very helpful to reduce start-up costs in ACP-SGD, fusing all tensors for one communication requires waiting the completion of gradient computation of BP, which will lose the opportunity of WFBP to hide communication overheads. 

To enable both WFBP and TF, it is suggested to merge a subset of compressed tensors of nearby layers. For example, given a three-layer DNN, during the BP from layer $3$ to $1$, one can merge and communicate the compressed gradients of layer $3$ and $2$ immediately when they are available, so that the fused communication task can be overlapped with the 
subsequent gradient computation task of layer $1$. 

\textbf{Buffer Size.} It is non-trivial to determine which tensors to be merged in order to minimize the communication overhead for different model and hardware configurations. In this work, we choose to allocate the buffer with a pre-defined buffer size, select the available tensors to fit in the buffer, and execute the all-reduce operation on the buffer. This has shown to be very effective in S-SGD~\cite{sergeev2018horovod,Li2020PyTorchDDP}. For example, the default buffer size in PyTorch-DDP is 25MB~\cite{Li2020PyTorchDDP}, and it will batch the gradient tensors of a ResNet-50 model (with $97.5$MB parameters) into $4$ buffers. The selected tensors are copied to one buffer, and the buffer is aggregated as a whole when it is ready. 

The buffer size is critical to control the trade-off between WFBP and TF. When the buffer size is too small, each tensor is communicated immediately to maximize the overlap (optimal WFBP), but it loses any opportunity of fusing tensors (no TF). When the buffer size is too large, all tensors will be merged together to be communicated at the end of BP (optimal TF), losing any change for overlapping (no WFBP). 

However, the compressed tensors ($P$ and $Q$) to be communicated in ACP-SGD are smaller than the gradient tensors ($M$), as shown in Fig.~\ref{fig:cdf-param}. This means the default buffer size (i.e., 25MB) used in S-SGD is not suitable for ACP-SGD anymore. For example, a ResNet-50 involves only $0.63$MB and $1.04$MB parameters in $P$ and $Q$, respectively, for ACP-SGD with a rank of $4$. To reduce the trouble of tuning the hyper-parameter of buffer size for different models and ranks, we choose to configure the compressed buffer size by scaling the default buffer size with the compression rate. For instance, the compression rates of ACP-SGD in ResNet-50 are $0.64\%$ and $1.07\%$ for $P$ and $Q$, giving compressed buffer sizes of $0.16$MB and $0.27$MB, respectively. Thus, it will batch $P$ tensors into 4 buffers, and batch $Q$ tensors into another 4 buffers for aggregation. Note that fusion results for $P$ and $Q$ are different to each other. \change{In this work, we use the default buffer size of 25MB, and the compressed buffer sizes are derived based on the compression rates. We notice that buffer size can be automatically tuned using e.g. Bayesian optimization technique~\cite{zhang2023decoupling}, but we do not adopt it in this work as the default buffer size can generally provide nearly optimal performance as studied later in Fig.~\ref{fig:effect-buffer-size}. }

\subsection{Implementation of ACP-SGD}
We implement our ACP-SGD prototype atop PyTorch. It wraps the SGD optimizer to cope with the underling gradient compression and communication operations. We use native APIs, i.e., torch.linalg.qr and torch.distributed.all\_reduce, to perform reduced QR decomposition for efficient orthogonalization (which is required for compression), and ring all-reduce for compressed gradient aggregation. The vector-shaped parameters (e.g., biases) requires no compression, while other parameters are reshaped into matrices for compression. 

To support WFBP, we register a hook function for each learnable parameter tensor during the back-propagation, and the registered hook function will be called when each gradient is ready. In the hook function, we implement the ACP-SGD with EF to compress the gradient $M$ into $P$ and $Q$ alternately. To support TF, the compressed tensors are copied into one of the buffers, and the ready buffer will be aggregated asynchronously via the all-reduce operation. At the end of back-propagation, we synchronize the all-reduce operations on the $P$ (or $Q$) buffers. 

\section{Evaluation}

\subsection{Experimental Settings}

For convergence experiments, we compare the performance of ACP-SGD with S-SGD and Power-SGD. We choose VGG-16~\cite{simonyan2014very} and ResNet-18~\cite{he2016deep} models on the Cifar-10~\cite{krizhevsky2009cifar} dataset. We run each algorithm for $300$ epochs on $4$ GPUs. We use per-GPU batch size of $128$ and learning rate of $0.1$, with a gradual warmup in the first $5$ epochs, and multiple learning rate decays (by a factor of $10$) at $150$ and $220$ epochs~\cite{goyal2017accurate}. Each algorithm uses a momentum of $0.9$, and Power-SGD and ACP-SGD use a rank of $4$. 

For time efficiency related experiments, we compare the iteration time of ACP-SGD with S-SGD and two versions of Power-SGD. The Power-SGD is the original implementation of~\cite{vogels2019powersgd}, which packs gradients after back-propagation for compression and communication. The Power-SGD with WFBP and TF optimizations (i.e., Power-SGD*) is implemented on PyTorch's communication hook~\cite{powerSGDhook}, which overlaps gradient compression and communication tasks with back-propagation. For a fair comparison, two Power-SGD baselines use reduced QR decomposition for orthogonalization like ACP-SGD. Note that S-SGD, Power-SGD*, and ACP-SGD are all optimized in the system-level with WFBP and TF, using a buffer size of $25$MB. We do not compare with other gradient compression methods, as we have shown that Power-SGD performed better than them in all cases (see \S\ref{sec:benckmark-exp}). For other model and testbed settings, we follow the same configurations in \S\ref{sec:benckmark-exp-setting}. 

\subsection{Convergence Verification}

\begin{figure}[!ht]
    \centering
    \includegraphics[width=0.45\linewidth]{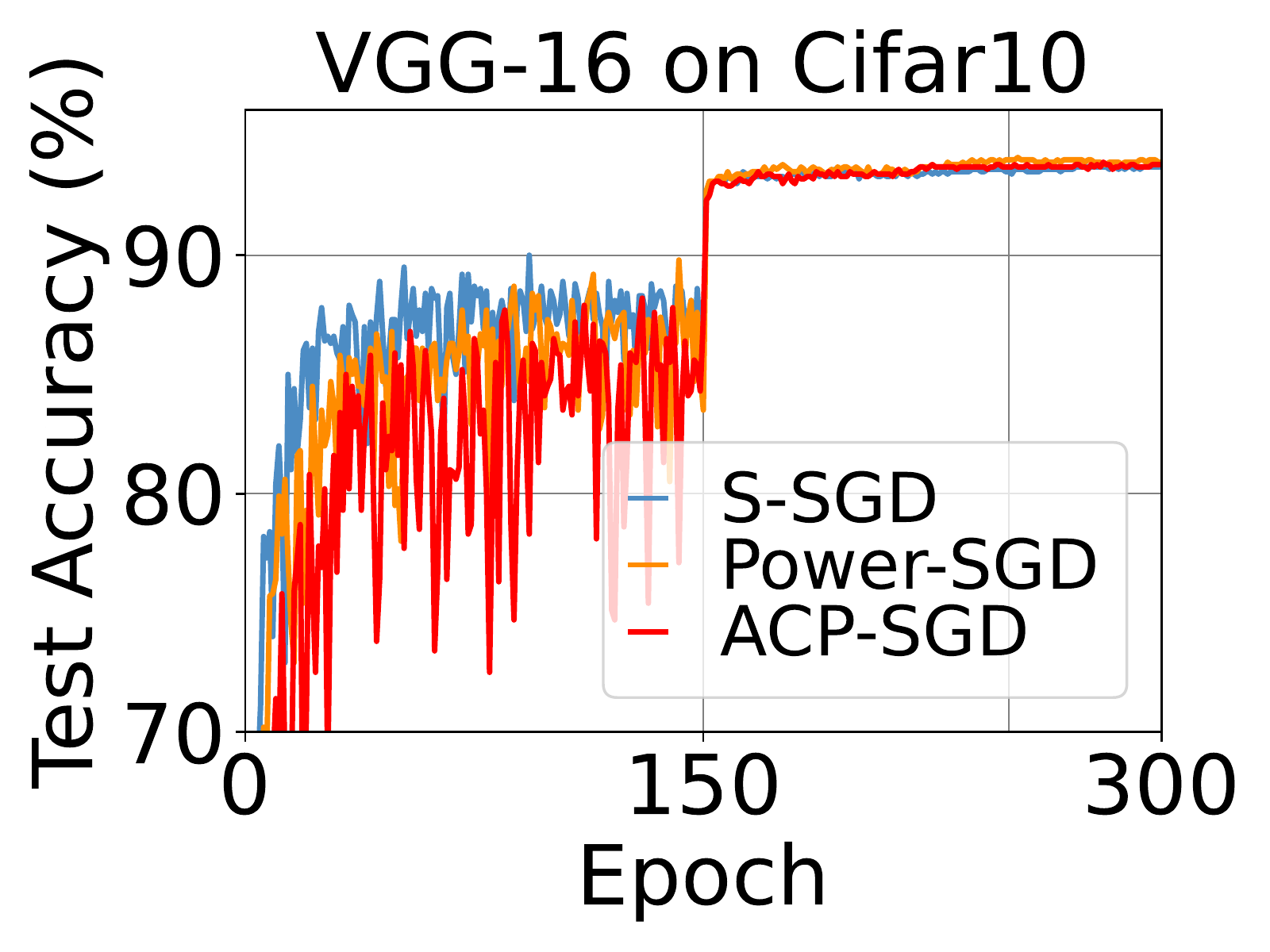}
    \includegraphics[width=0.45\linewidth]{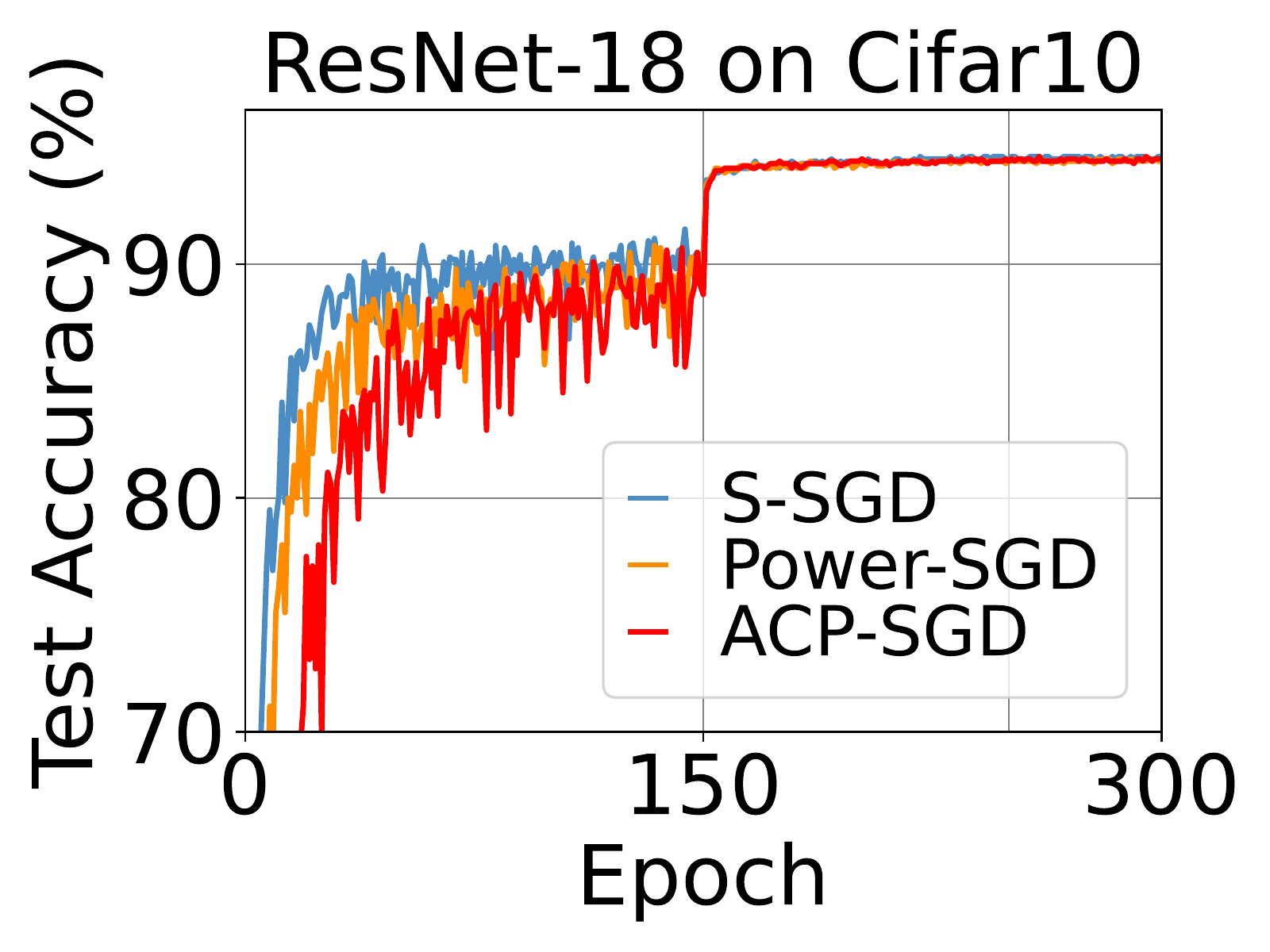}
    \caption{Convergence comparison among S-SGD, Power-SGD, and ACP-SGD on training VGG-16 and ResNet-18 models. }
    \label{fig:convergence}
\end{figure}

We first verify the convergence performance of our ACP-SGD on training two different models VGG-16 and ResNet-18 on the Cifar-10 dataset. The results are given in Fig.~\ref{fig:convergence}, showing that ACP-SGD can achieve very close accuracy results (i.e., $94.1\%$ for VGG-16,
and $94.6\%$ for ResNet-18) compared with S-SGD and Power-SGD. We see that gradient compression methods Power-SGD and ACP-SGD converge slightly slower than S-SGD in the early training stage, which however does not affect their final achieved model accuracy. 

\begin{figure}[!ht]
    \centering
    \includegraphics[width=0.45\linewidth]{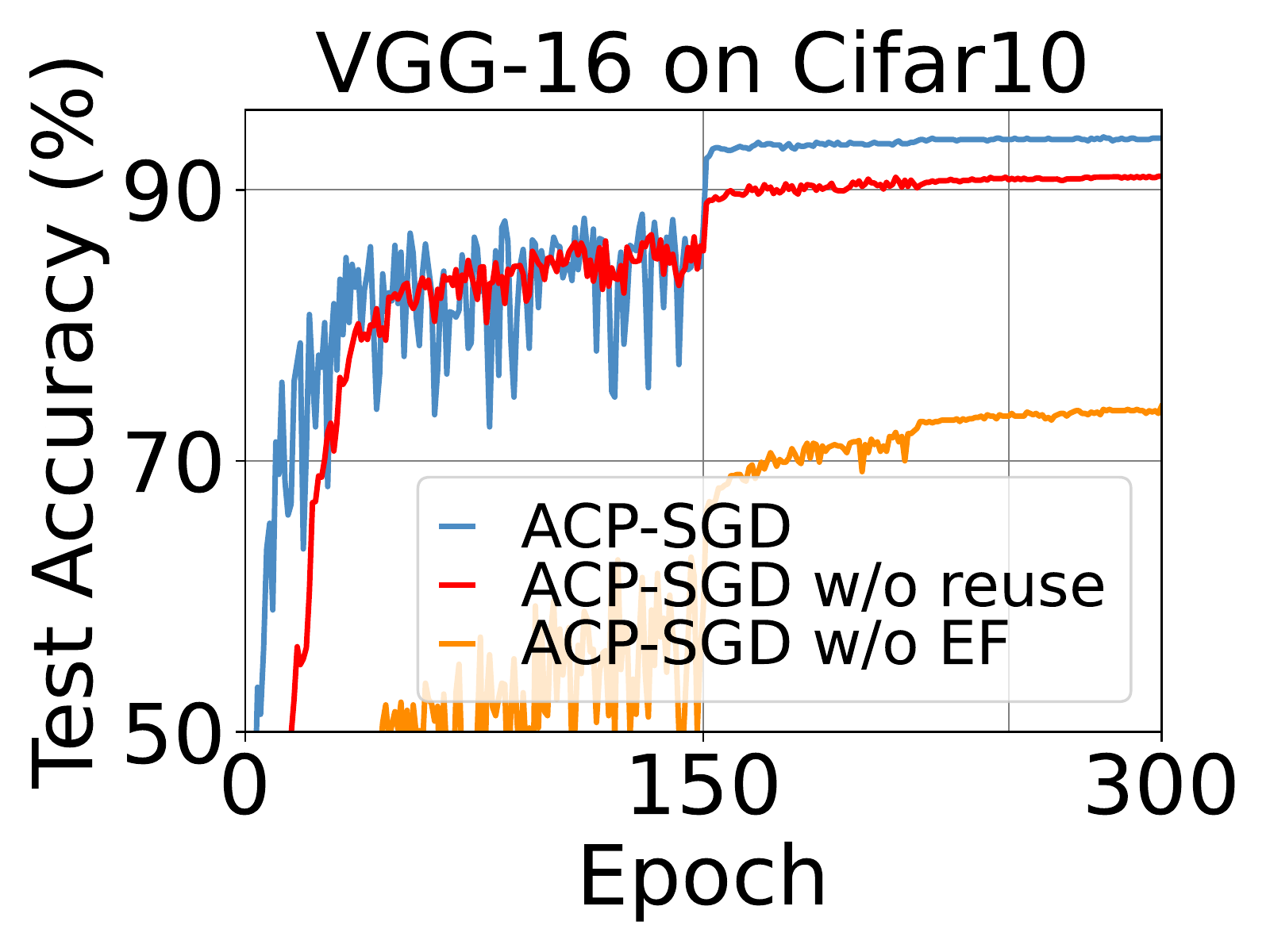}
    \includegraphics[width=0.45\linewidth]{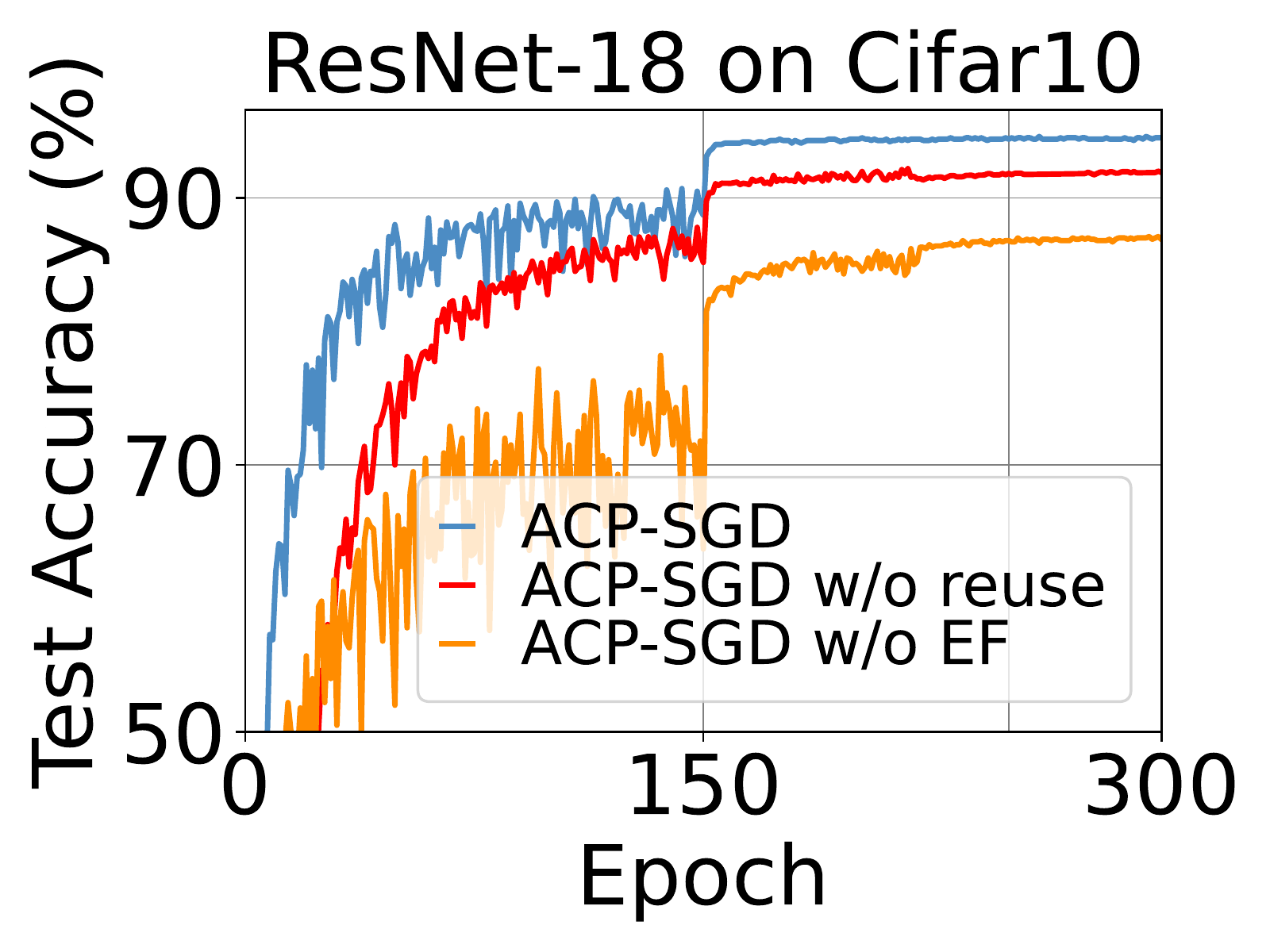}
    \caption{Convergence of ACP-SGD without error-feedback (EF) or reuse on training VGG-16 and ResNet-18 models. }
    \label{fig:convergence-ablation}
\end{figure}

We contribute the good convergence performance of ACP-SGD to the usage of EF and query reuse mechanisms. To validate it, we perform an ablation study by disabling one of them, and report their results in Fig.~\ref{fig:convergence-ablation}. It demonstrates that ACP-SGD without EF and reuse perform poorly, verifying that both of them are key components to ACP-SGD. 

\subsection{Wall-clock Iteration Time}
We compare the performance of our ACP-SGD with S-SGD~\cite{Li2020PyTorchDDP}, Power-SGD~\cite{vogels2019powersgd}, and Power-SGD* (i.e., Power-SGD optimized with WFBP and TF)~\cite{powerSGDhook}. We report the iteration time (mean$\pm$std) in Table~\ref{tab:acp-sgd-time}, where the best results are in bold. 

\begin{table}[!ht]
    \centering
    \begin{tabular}{|c|c|c|c|c|}
        \hline
        Model & S-SGD & Power-SGD & Power-SGD* & ACP-SGD \\\hline\hline
        ResNet-50 & $266\pm3$ & $302\pm4$ & $286\pm2$ & $\mathbf{248}\pm1$ \\\hline
        ResNet-152 & $500\pm1$ & $423\pm4$ & $404\pm4$ & $\mathbf{316}\pm3$ \\\hline
        BERT-Base & $805\pm4$ & $236\pm4$ & $292\pm4$ & $\mathbf{193}\pm1$ \\\hline
        BERT-Large & $2307\pm12$ & $392\pm1$ & $516\pm9$ & $\mathbf{245}\pm3$ \\\hline
    \end{tabular}
    \caption{Average iteration time (in milliseconds) comparison of ACP-SGD and other methods. }
    \label{tab:acp-sgd-time}
\end{table}

From Table~\ref{tab:acp-sgd-time}, it is observed that ACP-SGD consistently outperforms other methods in all tested cases. Specifically, ACP-SGD achieves an average of $4.06\times$, $1.34\times$, and $1.51\times$ speedups over S-SGD, Power-SGD, and Power-SGD* methods, respectively. And ACP-SGD can provide a significant speedup over S-SGD, e.g., training BERT-Large up to $9.42\times$ faster than S-SGD. However, two gradient compression baselines (Power-SGD and Power-SGD*) do not guarantee improved performance over S-SGD, for example, they run about $11\%$ slower than S-SGD in training ResNet-50. 

For Power-SGD* with system optimizations, it performs better than Power-SGD in small models (ResNet-50 and ResNet-152), but performs worse than Power-SGD in large models (BERT-Base and BERT-Large). This is because larger models typically require more compute-intensive gradient computation and compression tasks, and hence cause more severe performance interference for Power-SGD*. In contrast, ACP-SGD with system optimizations will not lead to resource competition, which in turns can achieve $1.60\times$ speedup than Power-SGD when training the BERT-Large model. The effects of system optimizations for Power-SGD and ACP-SGD are studied later in Fig.~\ref{fig:ablation-opt}. 

\begin{figure}[!ht]
    \centering
    \includegraphics[width=0.85\columnwidth]{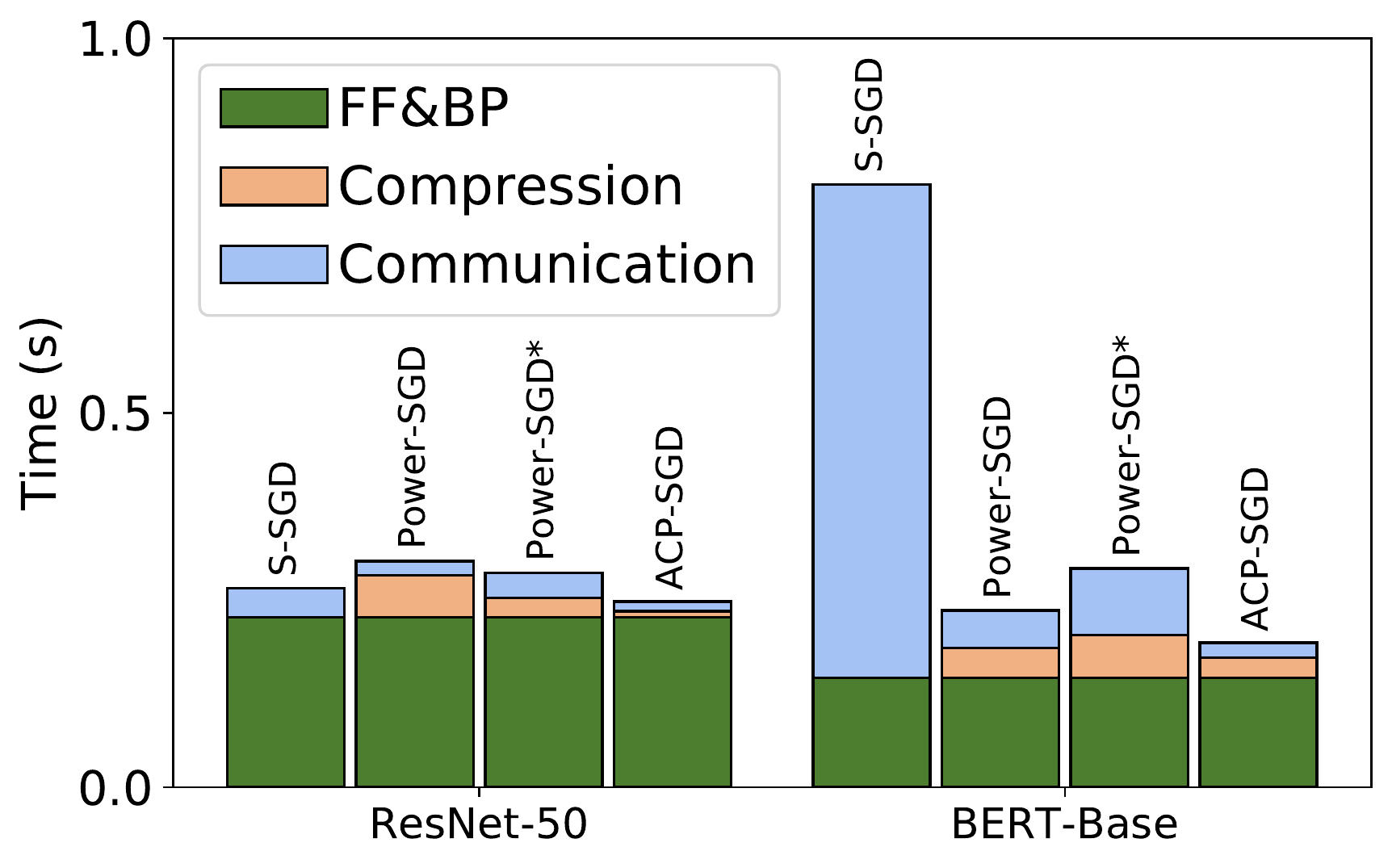}
    \caption{Time breakdowns of ACP-SGD and other methods on ResNet-50 and BERT-Base models.}
    \label{fig:acpsgd-breakdown}
\end{figure}

To further investigate the performance of ACP-SGD, we give the time breakdowns in Fig.~\ref{fig:acpsgd-breakdown} for two representative models (ResNet-50 and BERT-Base). It shows that ACP-SGD has very low gradient compression and communication overheads, and it achieves almost a linear scalability. Meanwhile, S-SGD can hide communication overhead very well for the small ResNet-50 model, but requires very high non-overlapped communication overhead for the large BERT-Base model. Power-SGD and Power-SGD* are useful to improve the training performance in large models, and our ACP-SGD moves a further step to improve the performance significantly.

\subsection{Benefits of System Optimizations}
\begin{figure}[!ht]
    \centering
    \subfloat[ResNet-152]{
        \includegraphics[width=0.45\columnwidth]{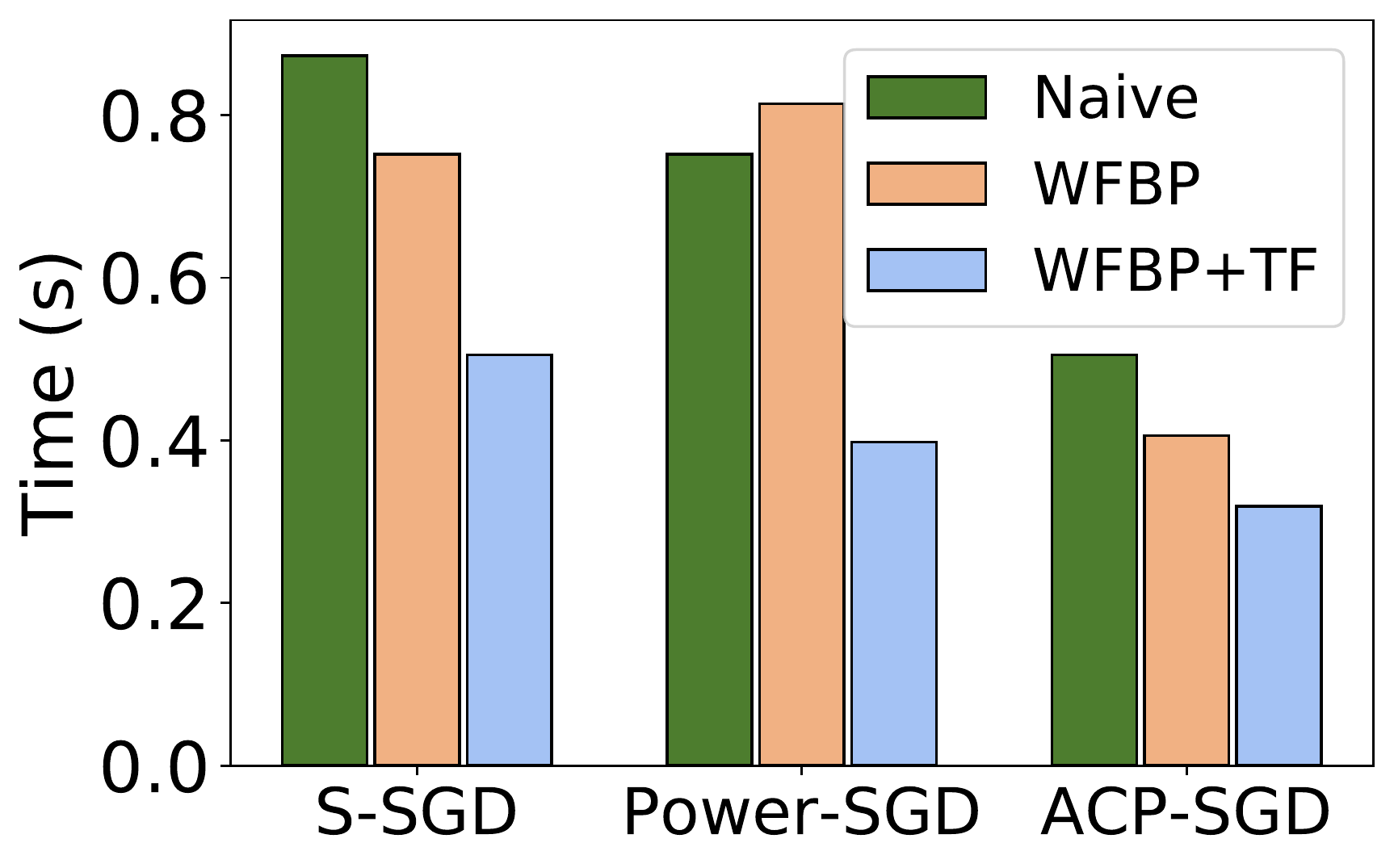}
    }
    \subfloat[BERT-Large]{
        \includegraphics[width=0.45\columnwidth]{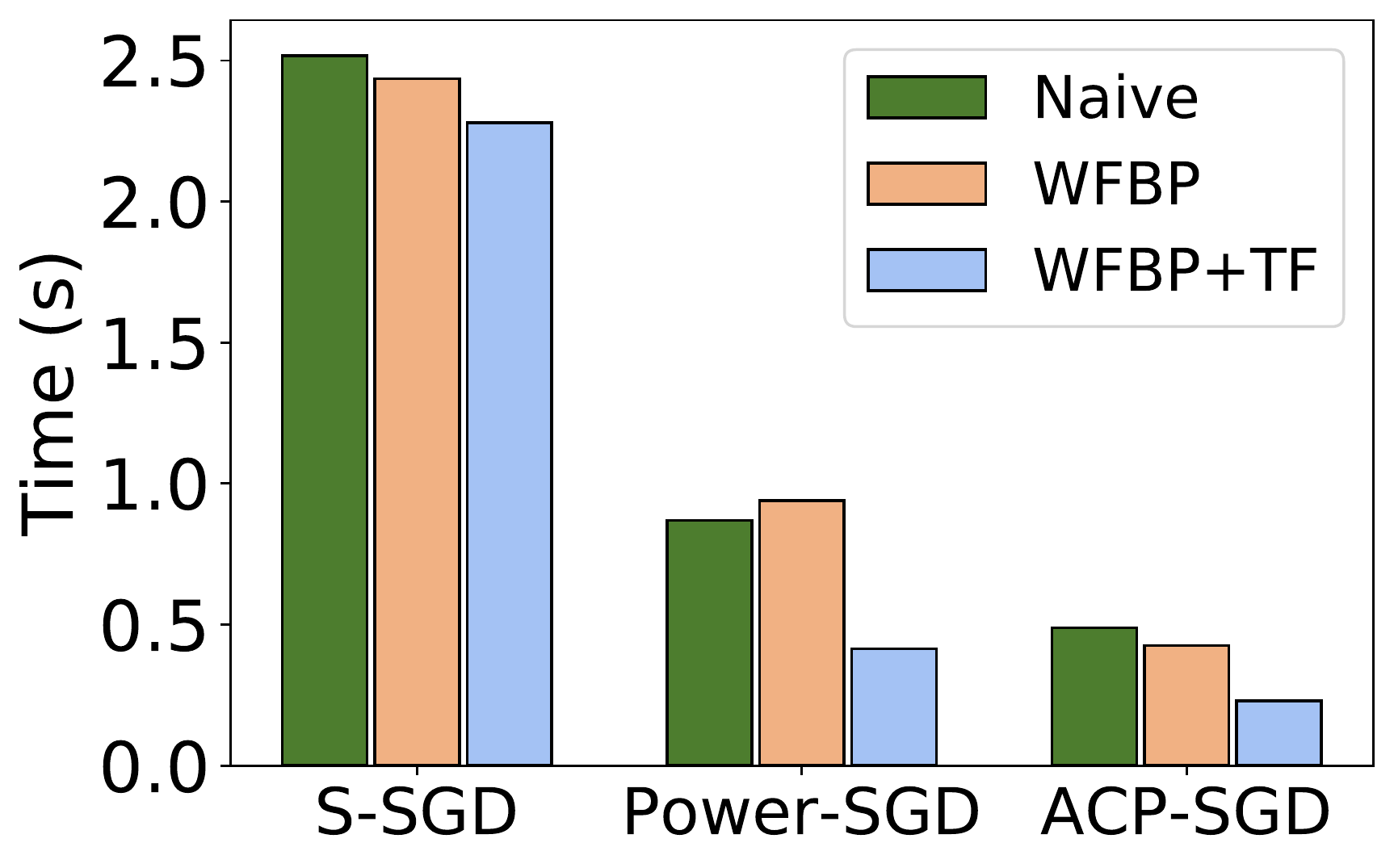}
    }
    \caption{Benefits of system optimizations, including wait-free back-propagation (WFBP) and tensor fusion (TF) techniques, for S-SGD, Power-SGD, and ACP-SGD methods. }
    \label{fig:ablation-opt}
\end{figure}

Next, we study the benefits of system optimizations (WFBP and TF) step-by-step for S-SGD, Power-SGD, and ACP-SGD. In the rest of the paper, Power-SGD refers to the implementation of Power-SGD* if it is not specified. We provide three variants for each method: Naive implementation without WFBP and TF, the one with only WFBP, and the one with WFBP and TF. We conduct experiments on two relatively large models ResNet-152 and BERT-Large, as they involve more communication optimizations than ResNet-50 and BERT-Base, respectively. The results are given in Fig.~\ref{fig:ablation-opt}. 

For S-SGD and ACP-SGD, it is observed that WFBP achieves about $12\%$ improvement over their Naive implementations. However, Power-SGD is not friendly to WFBP, since overlapping gradient computation and gradient compression causes compute resource competition, leading to an average of $13\%$ slowdown over its Naive implementation. Besides, we find that TF is able to provide a significant speedup for any method with WFBP. Specifically, WFBP with TF achieves an average of $1.28\times$, $2.16\times$ and $1.56\times$ speedups than WFBP (without TF) for S-SGD, Power-SGD, and ACP-SGD, respectively. In particular, TF for Power-SGD achieves the most significant improvement among three methods, due to the collective effect of reduced start-up cost and alleviated performance interference. As for ACP-SGD, with the help of WFBP and TF, it can achieve up to $2.14\times$ speedup over its Naive implementation, which validates the importance of two system optimizations for our gradient compression algorithm. 

\textbf{Sensitivity study of buffer size.} 
When using two system optimizations, buffer size is critical to control the trade-off between WFBP and TF. To study the effect of buffer size, we change the buffer size from $0$ to $1500$MB when training the BERT-Large model (with $1282.6$MB parameters), that is, from optimal WFBP without TF to optimal TF without WFBP. We perform Power-SGD and ACP-SGD with two different ranks: 32 and 256, where rank 256 (with $5.4\times$ compression ratio) is suggested to avoid possible performance degradation in~\cite{ramesh2021zero}.  

\begin{figure}[!ht]
    \centering
    \subfloat[Rank=32]{
        \includegraphics[width=0.45\columnwidth]{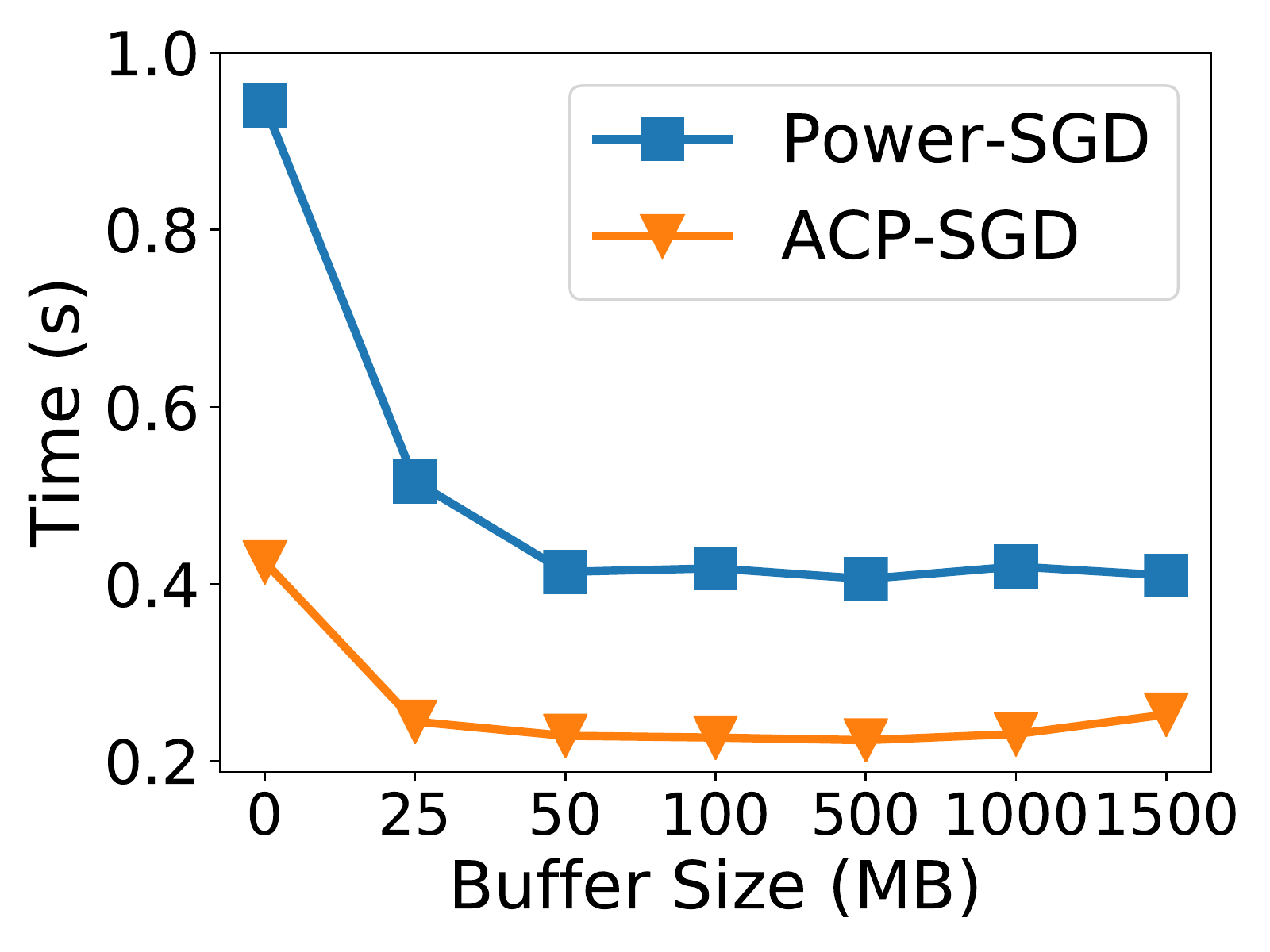}
    }
    \subfloat[Rank=256]{
        \includegraphics[width=0.45\columnwidth]{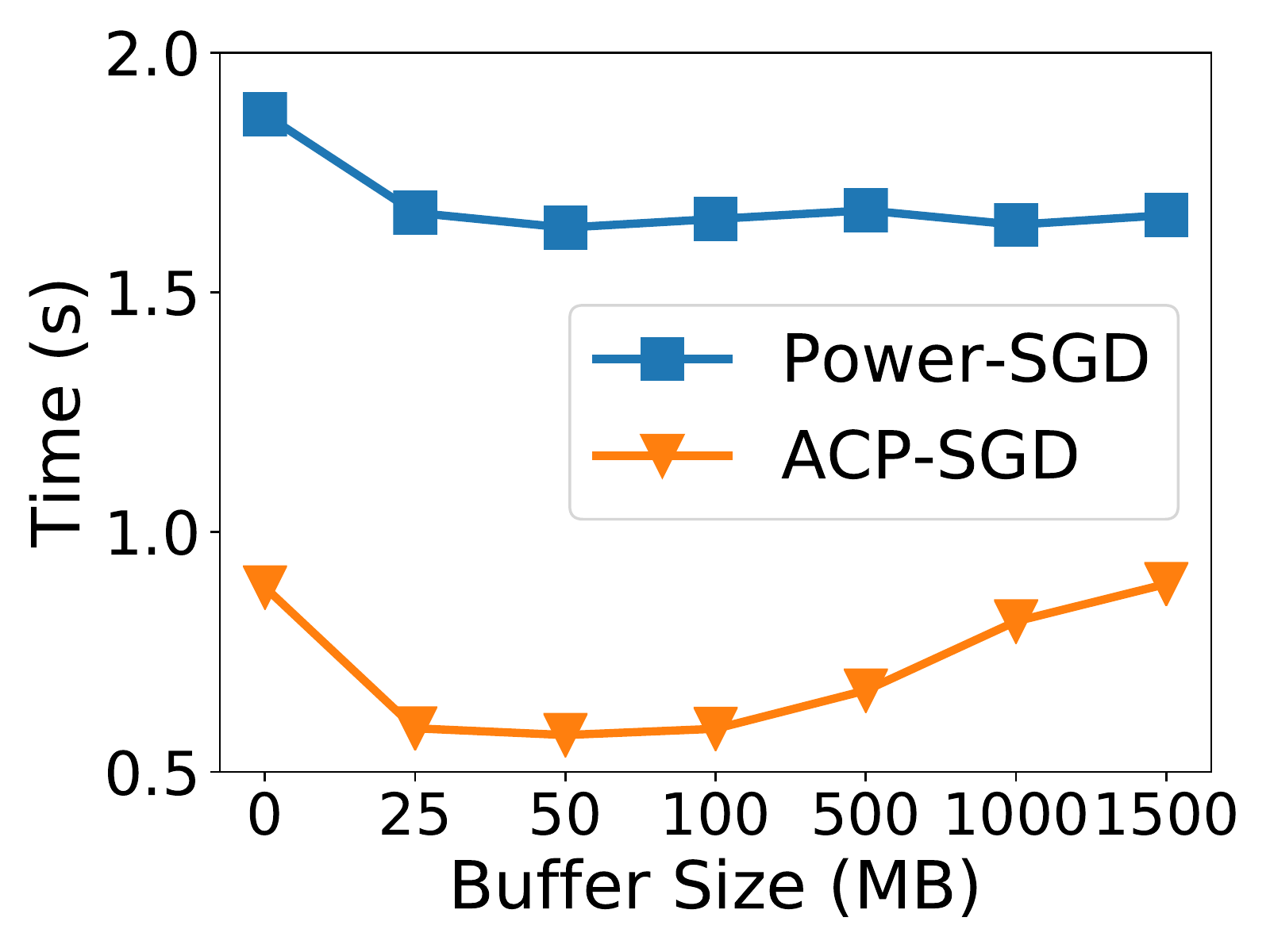}
    }
    \caption{Effect of buffer size in Power-SGD and ACP-SGD methods with different ranks when training BERT-Large. The default buffer size is 25MB. }
    \label{fig:effect-buffer-size}
\end{figure}

From Fig.~\ref{fig:effect-buffer-size}, we see that ACP-SGD can consistently outperform Power-SGD under different buffer size and rank settings. And ACP-SGD is very robust to the value of buffer size under different ranks, for example, the default buffer size of $25$MB remains a good candidate for ACP-SGD under different ranks (i.e., compression ratios). In particular, for rank=256, ACP-SGD with a buffer size of 25MB significantly outperforms two extreme cases: buffer size of 0MB (no TF) and buffer size of 1500MB (full TF), providing about $50\%$ improvement than both cases. We contribute it to that ACP-SGD can adaptively adjust the compressed buffer size for TF with different ranks, which can largely reduce the trouble for hyper-parameter tuning. 

\subsection{Effect of Hyper-parameters}
\begin{figure}[!t]
    \centering
    \subfloat[Batch size]{
        \includegraphics[width=0.47\columnwidth]{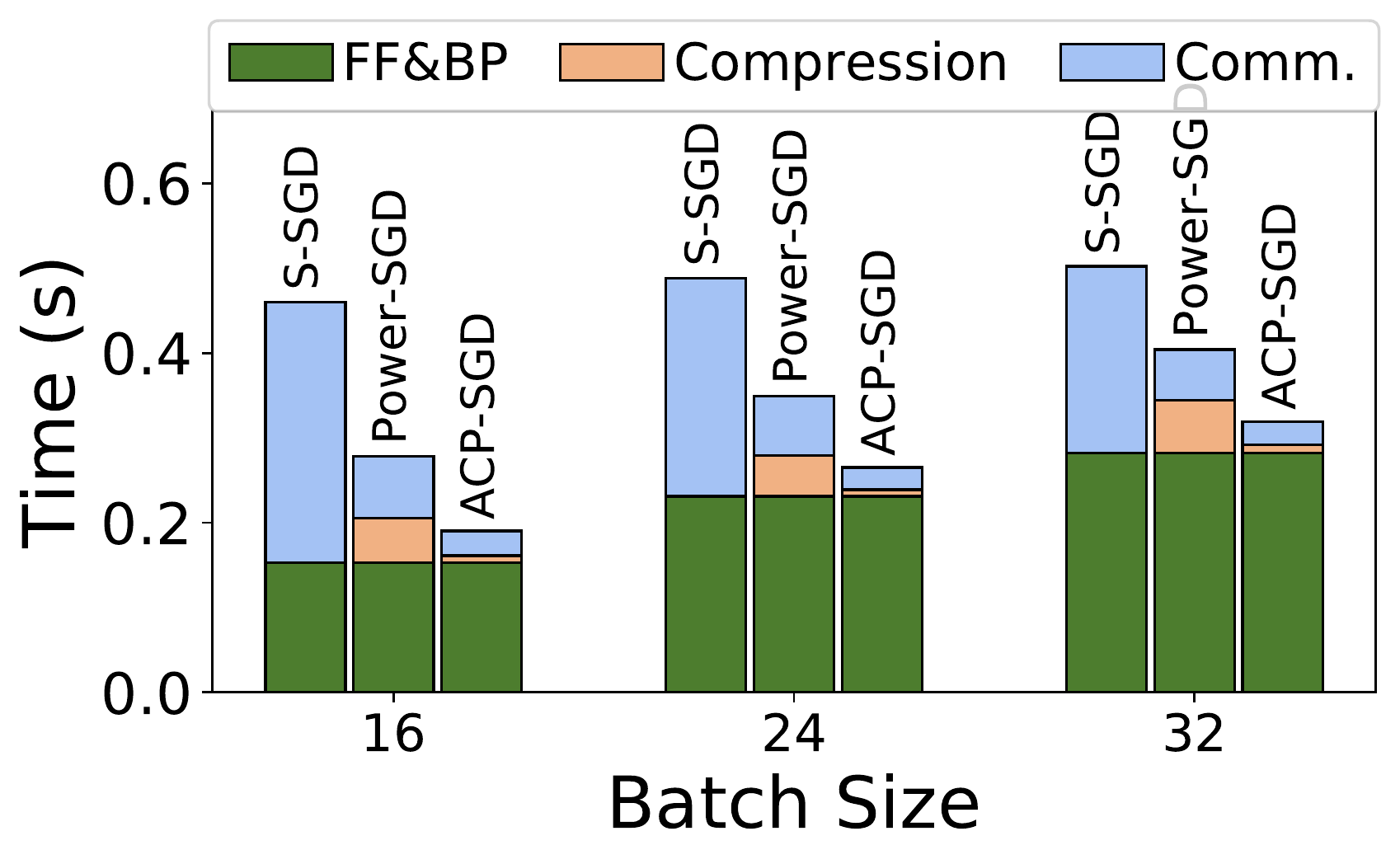}
    }
    \subfloat[Rank size]{
        \includegraphics[width=0.47\columnwidth]{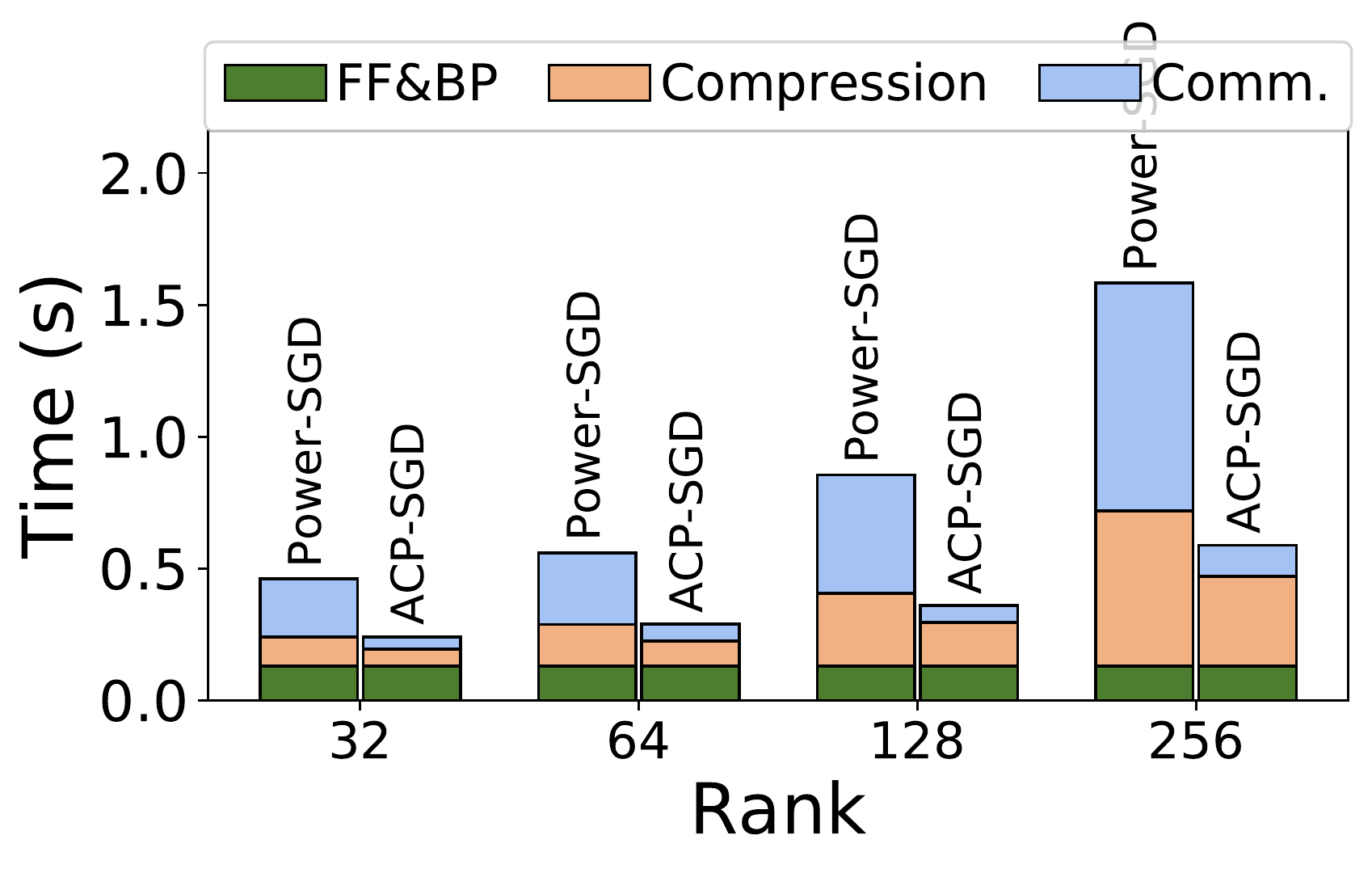}
    }
    \caption{Effect of hyper-parameters: (a) varying batch size for ResNet-152, and (b) varying rank size for BERT-Large. }\label{fig:effect-batch-size}\label{fig:effect-rank}
    \vspace{-8pt}
\end{figure}

\textbf{Effect of batch size. } Here we compare ACP-SGD with S-SGD and Power-SGD on different batch sizes for training the ResNet-152 model. We vary the per-GPU batch size from 16 to 32 to fully utilize the GPU memory, and keep other configurations the same. We report results in Fig.~\ref{fig:effect-batch-size}(a). We find that ACP-SGD consistently outperforms S-SGD and Power-SGD under different batch sizes. For example, when using batch size 16, ACP-SGD provides $2.4\times$ and $1.5\times$ speedups than S-SGD and Power-SGD, respectively. Using batch size 32, it gives $1.6\times$ and $1.3\times$ speedups over S-SGD and Power-SGD, respectively. ACP-SGD achieves the best performance since it always has very small compression and communication costs for different batch sizes. Besides, for the three methods, using large batch sizes often provides performance improvement in terms of throughput. For instance, S-SGD, Power-SGD, and ACP-SGD achieve $3.7\times$, $2.8\times$, $2.4\times$ speedups on throughput by increasing batch size from $16$ to $32$. Especially for S-SGD (with the heaviest communication), we observe that its non-overlapped communication overhead drops as the batch size increases. This is because increasing batch size leads to an increase in the computation-to-communication ratio, which in turns benefits S-SGD to overlap more communications with computations. Using larger batch sizes may further reduce the performance gap between S-SGD and ACP-SGD, but it is impractical due to the limited GPU memory capacity. 

\textbf{Effect of rank. } The parameter of rank is used to control the compression ratio of Power-SGD and ACP-SGD. The lower the rank, the stronger the compression. To show the effect of rank, we vary the rank from 32 to 256 by a factor of 2 for training BERT-Large (with model dimension of 1024). We report the performance under different ranks in Fig.~\ref{fig:effect-rank}(b), showing that using large ranks leads to an increase in the compression and communication overheads for both Power-SGD and ACP-SGD. Therefore, Power-SGD and ACP-SGD have $3.4\times$ and $2.4\times$ higher total iteration time, respectively, when varying the rank from 32 to 256. However, compared to Power-SGD, ACP-SGD can overlap more communication overheads with gradient computation and compression overheads as the rank increases. For example, ACP-SGD provides $1.9\times$ speedup than Power-SGD when training using rank 32, and this speedup becomes $2.7\times$ for rank 256. In particular, when using rank 256, ACP-SGD has almost $7.3\times$ reduction in the non-overlapped communication time over Power-SGD. In addition, for training BERT-Large, ACP-SGD with rank 256 (i.e., with $5.4\times$ compression ratio) can still achieve about $3.9\times$ improvement over S-SGD. 

\subsection{Effect of Cluster Settings}
\textbf{Effect of the number of GPUs.} To study the scalability of ACP-SGD, we vary number of GPUs from 8 GPUs (2 nodes) to 64 GPUs (16 nodes). The interconnect between nodes is 10GbE. We use ring all-reduce for gradient aggregation. 

\begin{figure}[!ht]
    \centering
    \subfloat[ResNet-50]{
        \includegraphics[width=0.45\columnwidth]{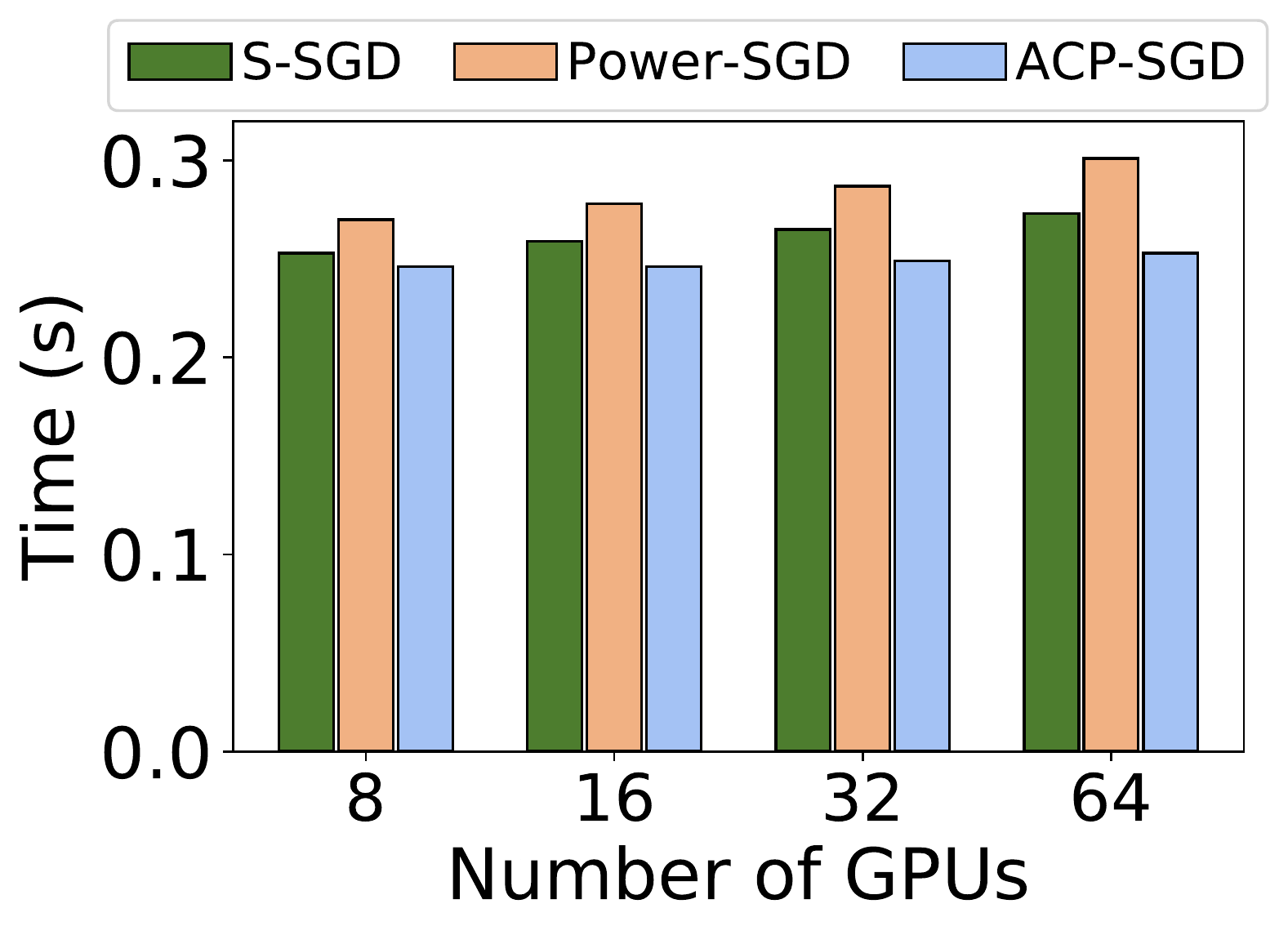}
    }
    \subfloat[BERT-Base]{
        \includegraphics[width=0.45\columnwidth]{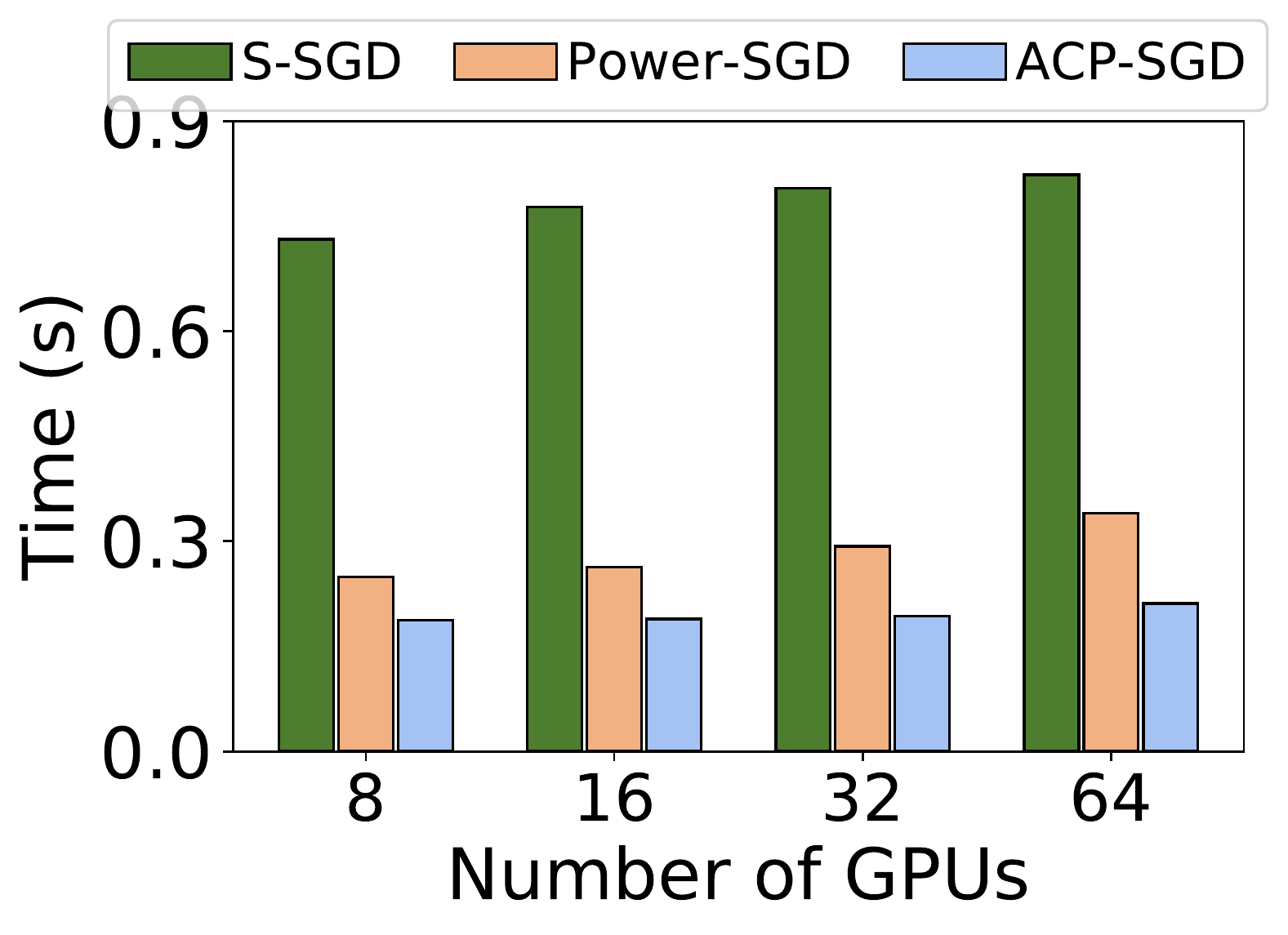}
    }
    \caption{Effect of varying the number of GPUs. }
    \label{fig:effect-ngpu}
    \vspace{-5pt}
\end{figure}

We report the effect of the number of GPUs in Fig.~\ref{fig:effect-ngpu}. It shows that all three methods have high scalability as the number of GPUs increases. For instance, it is observed that S-SGD, Power-SGD, and ACP-SGD have only an average of $10\%$, $24\%$, and $8\%$ increase in per iteration time, respectively, when scaling from $8$ GPUs to $64$ GPUs. They can scale well because of using 1) ring all-reduce primitive with optimal bandwidth, whose communication  complexity stays almost constant with increase in number of GPUs, and 2) tensor fusion to amortize start-up cost. While our testbed is up to 64 GPUs, the high scalability of ACP-SGD implies that it can always outperform S-SGD and Power-SGD using more GPUs. 

\begin{figure}[!ht]
    \centering
    \subfloat[ResNet-50]{
        \includegraphics[width=0.45\columnwidth]{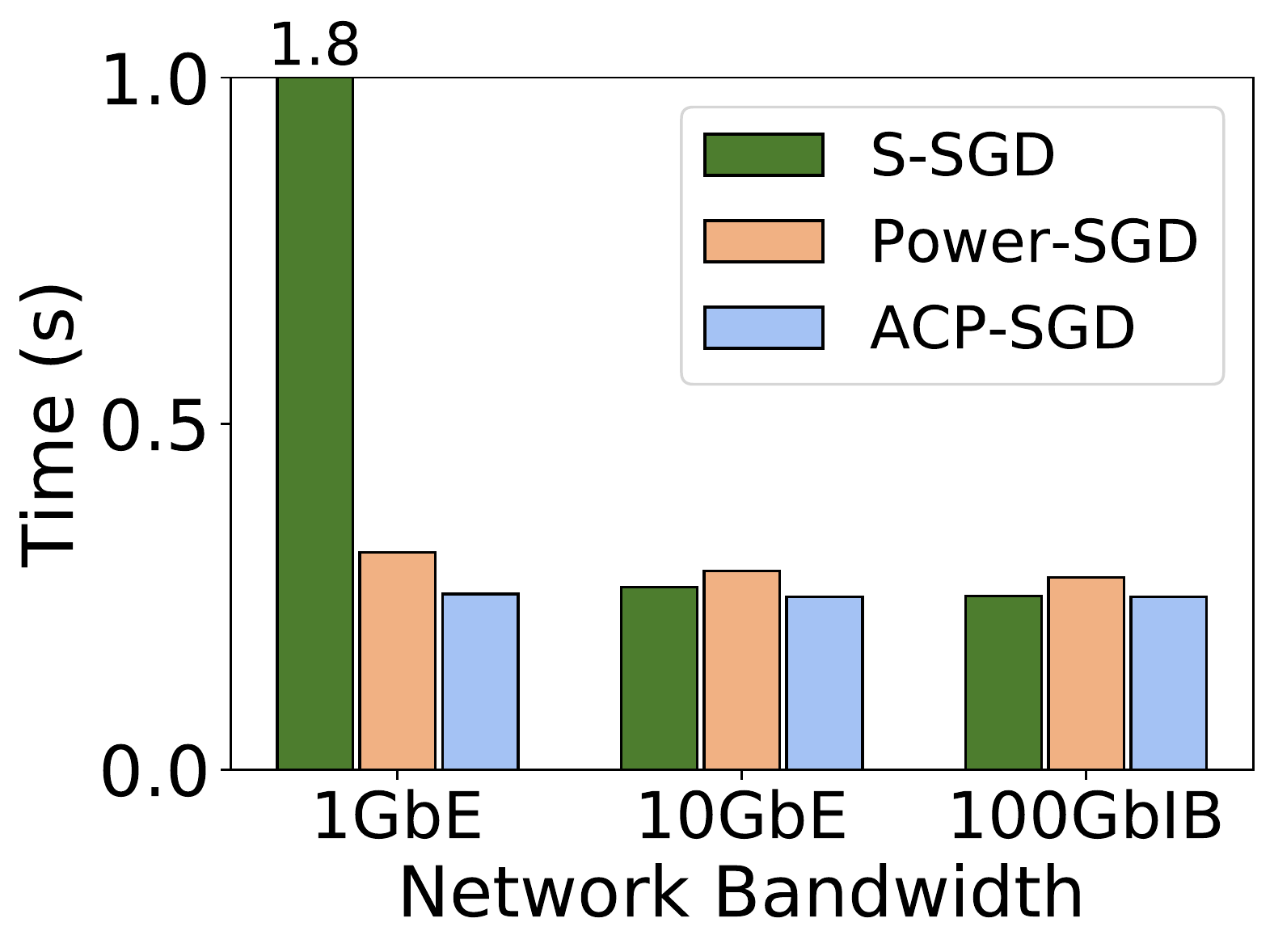}
    }
    \subfloat[BERT-Base]{
        \includegraphics[width=0.45\columnwidth]{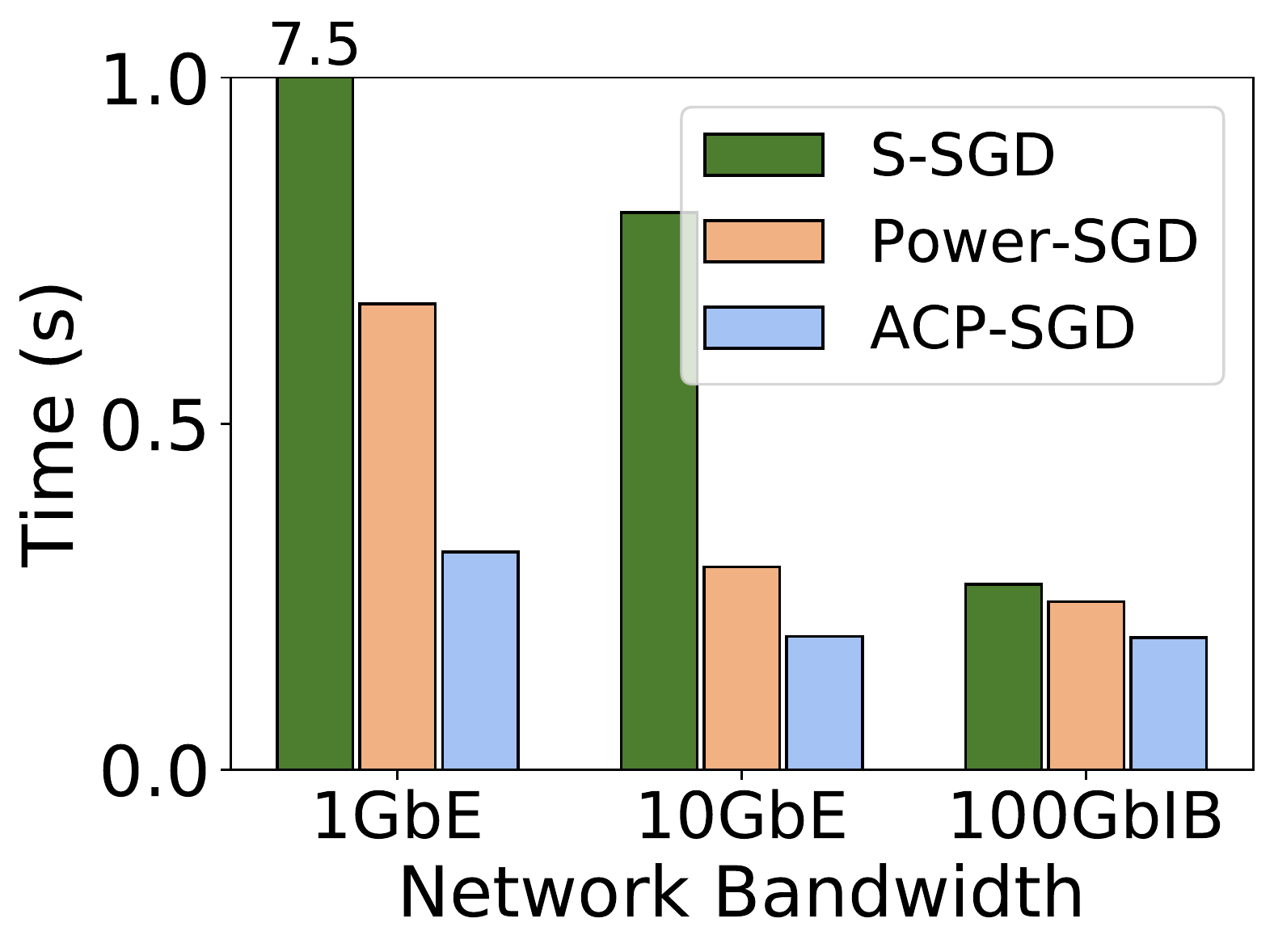}
    }
    \caption{Effect of network bandwidth. We display the iteration time on top of the bar when it goes beyond y-axis limit. }
    \label{fig:effect-bandwidth}
\end{figure}
\textbf{Effect of network bandwidth.} Here we validate the scalability of ACP-SGD on 32 GPUs, connected by different levels of networks, including inexpensive commodity 1Gb/s Ethernet (1GbE), ubiquitous data-center 10Gb/s Ethernet (10GbE), and expensive high-bandwidth 100Gb/s Infiniband (100GbIB). 

The results are given in Fig.~\ref{fig:effect-bandwidth}, showing that ACP-SGD performs better than S-SGD and Power-SGD under different networks. For slow 1GbE network, gradient compression methods Power-SGD and ACP-SGD can largely outperform S-SGD. In ResNet-50, Power-SGD and ACP-SGD achieves $5.7\times$ and $7.1\times$ speedups over S-SGD, and the speedups increase up to $11.2\times$ and $23.9\times$ in BERT-Base. For fast 100GbIB network, while the communication overhead of S-SGD can be well alleviated, ACP-SGD can provide about $40\%$ improvement over S-SGD when training BERT-Base. 


\section{Conclusion}
In this paper, we studied gradient compression methods that mitigate communication bottleneck in distributed deep learning. We first evaluated the efficacy of three representative gradient compression methods, and it was observed that current compression methods performed poorly in many cases. To optimize the performance of gradient compression, we then proposed the alternate compressed Power-SGD (ACP-SGD) algorithm, that can 1) achieve model accuracy on par with S-SGD using error feedback and reuse mechanisms, and 2) outperform the efficiency of S-SGD and Power-SGD with several system optimization techniques. Finally, we conducted extensive experiments to show that our ACP-SGD can consistently outperform other popular baselines across many setups.

\section*{Acknowledgments}
The research was supported in part by a RGC RIF grant under the contract R6021-20, RGC GRF grants under the contracts 16209120, 16200221 and 16207922, and the National Natural Science Foundation of China (NSFC) (Grant No. 62272122). 


\bibliographystyle{IEEEtran}
\bibliography{cites-short}

\end{document}